\documentclass[letterpaper, journal]{IEEEtran}
\usepackage{amsmath,amsfonts}
\usepackage{algorithmic}
\usepackage{algorithm}
\usepackage{array}
\usepackage[caption=false,font=normalsize,labelfont=sf,textfont=sf]{subfig}
\usepackage{textcomp}
\usepackage{stfloats}
\usepackage{url}
\usepackage{verbatim}
\usepackage{graphicx}
\usepackage{cite}
\usepackage{graphics} 
\usepackage{epsfig} 
\usepackage{mathptmx} 
\usepackage{times} 
\usepackage{amsmath} 
\usepackage{amssymb} 
\graphicspath{ {fig/} }
\usepackage[dvipsnames]{xcolor}
\usepackage{balance} 
\usepackage{hyperref}
\usepackage{float}
\usepackage{layouts}  
\usepackage{booktabs}
\usepackage{multirow}
\usepackage{fancyvrb}
\usepackage{fvextra}
\usepackage[shortlabels]{enumitem}
\usepackage[utf8]{inputenc} 
\DeclareUnicodeCharacter{2205}{\ensuremath{\emptyset}} 

\begin{document}

\title{Harnessing Language for Coordination: A Framework and Benchmark for LLM-Driven Multi-Agent Control}

\author{Timothée Anne$^{1}$, Noah Syrkis$^{1}$, Meriem Elhosni$^{2}$, Florian Turati$^{2}$, Franck Legendre$^{2}$, Alain Jaquier$^{2}$, and Sebastian Risi$^{1}$
\thanks{$^{1}$IT University of Copenhagen, Copenhagen, Denmark}
\thanks{$^{2}$armasuisse Science+Technology, Thun, Switzerland}
}



\maketitle

\begin{abstract}
Large Language Models (LLMs) have demonstrated remarkable performance across various tasks. Their potential to facilitate human coordination with many agents is a promising but largely under-explored area. Such capabilities would be helpful in disaster response, urban planning, and real-time strategy scenarios. In this work, we introduce (1) a real-time strategy game benchmark designed to evaluate these abilities and (2) a novel framework we term HIVE. HIVE empowers a single human to coordinate swarms of up to 2,000 agents through a natural language dialog with an LLM. We present promising results on this multi-agent benchmark, with our hybrid approach solving tasks such as coordinating agent movements, exploiting unit weaknesses, leveraging human annotations, and understanding terrain and strategic points. Our findings also highlight critical limitations of current models, including difficulties in processing spatial visual information and challenges in formulating long-term strategic plans. This work sheds light on the potential and limitations of LLMs in human-swarm coordination, paving the way for future research in this area. The HIVE project page, \url{hive.syrkis.com}, includes videos of the system in action.
\end{abstract}


\section{Introduction}

The growing capabilities of Large Language Models (LLMs) have opened up new frontiers in artificial intelligence, including enabling human-AI collaboration across diverse and complex domains \cite{sharma2024, bakhtin2022, dinan2022}. While much of the existing research focuses on LLMs' proficiency in tasks like natural language understanding and generation, their potential for coordinating is a new area of exploration~\cite{lore2024, zhang2024b}. This ability can be particularly critical in disaster response, urban planning, and strategy games, where efficient coordination of multiple agents can significantly impact the operation's outcome.

\begin{figure*}[t]
    \centering
    \includegraphics[width=0.95\linewidth]{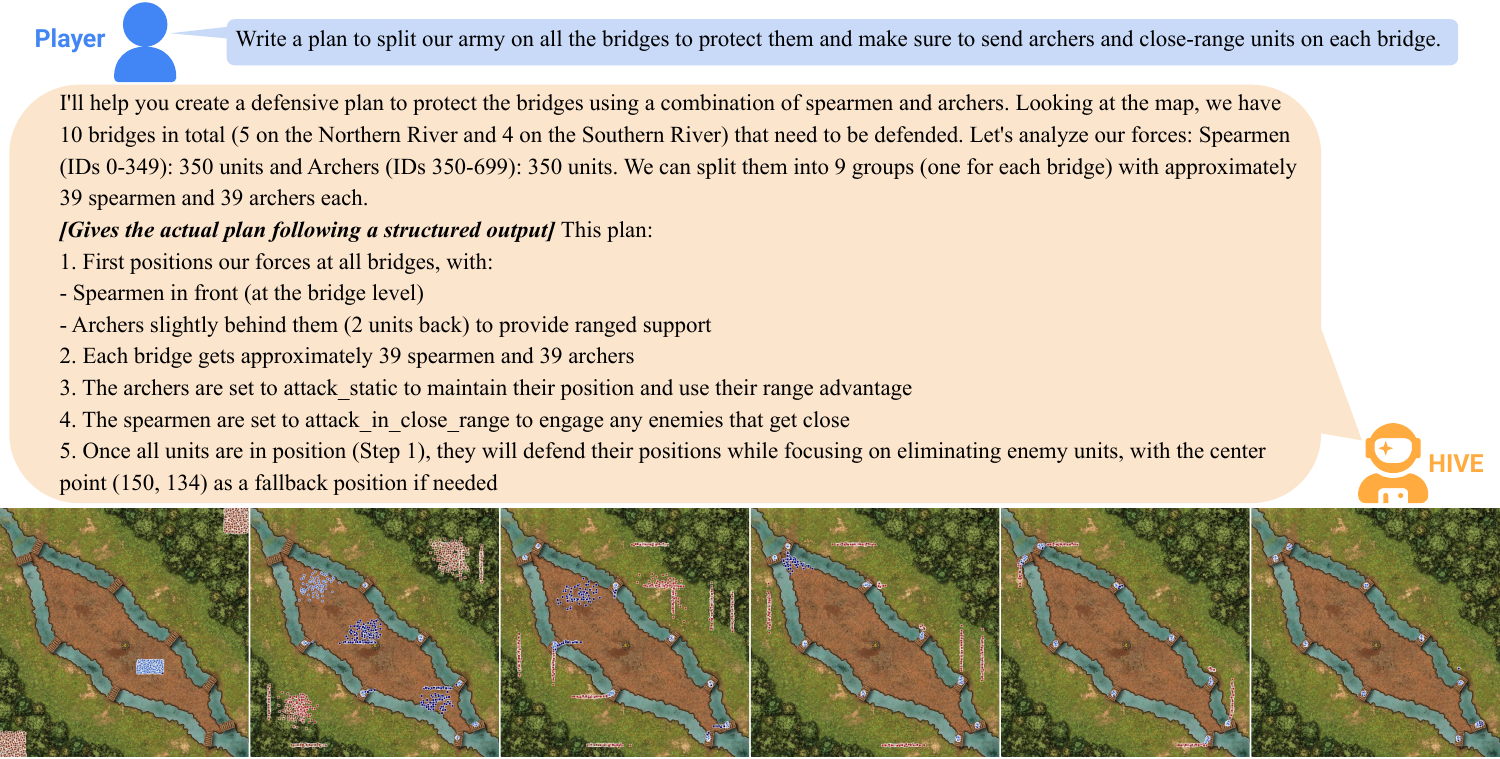}
    \caption{\textbf{Example of interactions between the player and  HIVE.} To win this scenario, the player and HIVE have to devise a plan to prevent the enemy units (in red) from reaching the center of the camp. To do so, the player proposes a high-level description, which HIVE takes as input to write down an actual plan using a structured output. In this example, HIVE also briefly describes the situation and the plan. Notice that even though the situational description is not always correct (ten bridges instead of nine), the actual plan is correct and wins this scenario. In this example, HIVE uses claude-3-5-sonnet-20241022 as the underlying LLM. Snapshots from the corresponding game are shown at the bottom.}
    \label{fig:teaser_figure}
\end{figure*}

This paper introduces a novel framework called \textbf{HIVE} (\emph{Hybrid Intelligence for Vast Engagements}) designed to facilitate such coordination by enabling natural language-based control of thousands of agents in real-time. Leveraging the strengths of LLMs, HIVE translates high-level human instructions into operational plans for agent swarms. Specifically, after receiving a high-level strategy prompt from the player, HIVE generates a plan following a structured output. This plan assigns each unit a target position and a behavior tree that takes an action at each step based on the unit's local observation. Fig.~\ref{fig:teaser_figure} illustrates an example where HIVE responds to the player’s prompt and devises a successful plan in a defense scenario.

The HIVE framework enables a new LLM-based benchmark that evaluates five core capabilities of multi-agent systems: coordination, exploitation of weaknesses, adherence to spatial markers, terrain utilization, and strategic planning (Sec.~\ref{sec:benchmark}). To our knowledge, it is the first real-time strategy game benchmark for LLMs in large-scale multi-agent systems. In this paper, we compared nine state-of-the-art LLMs.
Additionally, we conduct detailed evaluations to address the following questions: How does HIVE scale to a larger number of units? (Sec.~\ref{sec:llm_scaling}) Does HIVE benefit from machine-human collaboration? (Sec.~\ref{sec:llm_alone}) And, how does HIVE's performance scale with different input modalities, e.g., a visual vs. a text description of the map (Sec.~\ref{sec:LLM_with_image})?

Our findings demonstrate that HIVE can execute complex multi-agent tasks successfully by combining human strategic input with LLM-powered operational planning. However, the study also reveals the current limitations of LLMs, including their sensitivity to input variations, challenges in visual-spatial reasoning, and difficulties with long-term strategic planning. These insights underline the dual nature of LLMs as both powerful tools for enhancing human decision-making and as systems requiring significant refinement for seamless integration into dynamic, multi-agent environments. 

This work sets the stage for advancing human-AI collaboration in multi-agent coordination tasks, highlighting both the potential and the hurdles of deploying LLMs in such contexts. By addressing these challenges, future research should be able to refine these systems to unlock their full potential.

\section{Related Work}
Recent works have demonstrated the potential of LLMs and vision language models (VLMs) in complex strategic games. Cicero~\cite{dinan2022} showed that LLMs can achieve human-level performance in Diplomacy through natural language negotiation and tactical coordination. SwarmBrain~\cite{shao2024}, an agent for StarCraft II, combines high-level LLM-based strategic planning with low-level non-LLM-based tactical execution. These works demonstrate that LLMs can reason about and execute strategic decisions in game environments. However, controlling multiple units in RTS games presents unique challenges. One approach developed by Zhou et al.~\cite{zhou2021} is a hierarchical command and control architecture, tested on StarCraft, which combines high-level and low-level reinforcement learning agents to balance micro and macro management. 

Using behavior trees for low-level actions with LLMs has been explored in robotics, translating the behavior tree from one domain to another~\cite{cao2023}. However, they only focused on generating behavior trees and did not evaluate their execution.

Furthering the ability to conduct AI research in the RTS domain, the StarCraft Multi-Agent Challenge (SMAC) environment \cite{samvelyan2019} established key benchmarks for multi-agent control~\cite{rashid2020, yu2021a}. SMACv2~\cite{ellis2023} extends this work, introducing procedurally generated scenarios requiring generalizations. Recently, JaxMARL's SMAX~\cite{rutherford2023} has enabled easy parallelization in an SMAC-like environment, lowering the barrier of entry for research into AI applied in an RTS setting. However, while these frameworks provide foundations for developing multi-agent control systems, they focus on tactical execution rather than strategic planning, leaving the integration with higher-level decision-making under-explored.

Significant recent research has focused on LLM-based multi-agent systems. A comprehensive survey discussing key aspects of agent profiling, communication, and environment interaction has also been written~\cite{guo2024}. For example, AgentCoord~\cite{pan2024} is a visual interface designed for coordination strategies in multi-agent collaboration to solve a joint goal. Challenges in multi-agent LLM systems include effective task allocation, robust reasoning, and efficient memory management~\cite{han2024a}. A key limitation across these systems is the computational overhead of LLM inference, which makes real-time applications challenging without significant architectural compromises (partly motivating hybrid systems like SwarmBrain, in which low-level execution is done without the LLM).

Zhang et al.~\cite{zhang2024b} surveyed the strategic reasoning capabilities of LLMs, highlighting their ability to understand game structures and adapt to different contexts. Kramár et al.~\cite{kramar2022} investigated how artificial agents can use communication for better cooperation in strategic games (e.g., Diplomacy). Lorè and Heydari~\cite{lore2024} analyzed how different LLMs are affected by contextual framing and game structure.

However, in sufficiently complex and dynamic environments, LLMs and VLMs struggle to perform well~\cite{paglieri2024, ruoss2024lmact}, sometimes even performing worse when visual information like an image of the game map is included.  LLMs also struggle with physical common-sense reasoning in 3D environments, performing worse than human children on basic spatial reasoning tasks~\cite{mecattaf2024}. 
Furthermore, long-range reasoning problems are exacerbated because these models are prone to hallucinations~\cite{wang2024b}. Although algorithms have been proposed to fortify against hallucination in multi-step reasoning by dividing up reasoning steps, the problem of long-range causal reasoning remains unsolved~\cite{wang2024b}. This is particularly problematic in strategy games, where decisions must build on each other, and errors compound over time.
AndroidArene\cite{xing2024} is a benchmark and environment that evaluates an LLM's ability to navigate an operating system (e.g., using a smartphone with apps), showing this to be a current limitation of LLMs.

Models also struggle to generalize beyond their training data when moving toward geospatial modalities, prompting the need for dedicated Vision-Language Geo-Foundation models \cite{zhou2024a}. This limitation is relevant for strategy games involving spatial reasoning and map awareness, where current models often fail to develop coherent long-term strategies based on terrain and unit positioning (and sometimes struggle to identify basic landmarks). This echoes recent research that shows VLMs often struggle with trivial tasks (e.g., determining if two circles intersect) and a necessity for spatial reasoning \cite{rahmanzadehgervi2025}.

Integrating LLMs into real-time strategy games faces several challenges not fully addressed by current research.
This suggests that while LLMs can engage in high-level strategic planning, they need significant support systems to translate these plans into tactical actions.

\section{The HIVE approach}
\label{sec:method}
\begin{figure*}[t]
    \centering
    \includegraphics[width=\linewidth]{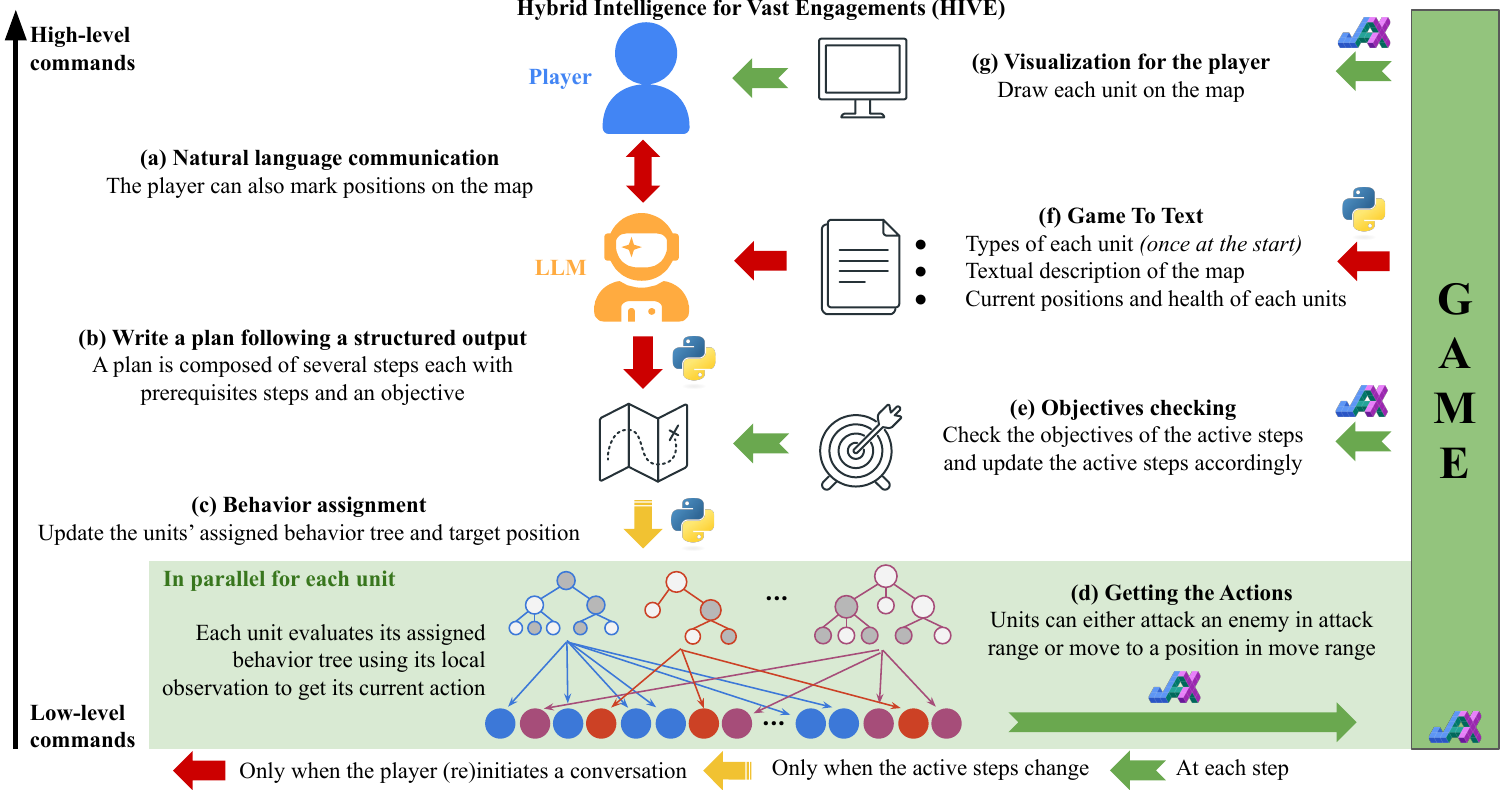}
    \caption{\textbf{Overview of the HIVE approach.} HIVE enables players to command up to several thousand units by delegating the tedious task of assigning behaviors and objectives to each unit to a general-purpose LLM. HIVE operates in two main phases: (\textbf{1}) a discussion phase with an LLM to develop a plan before the game starts and (\textbf{2}) the execution phase, where the plan assigns a behavior tree to each unit. We implemented the modules that handle the units in JAX~\cite{jax2018github} and those that manage high-level information in Python. See the main text for a more detailed description.}
    \label{fig:method_overview}
\end{figure*}

Our approach, HIVE, allows a player to control thousands of units in a strategy game through a natural language LLM-based dialog. We hypothesize that this type of human-machine collaboration should alleviate some of the aforementioned shortcomings of purely LLM-based approaches.

Fig.~\ref{fig:method_overview} presents an overview of its main components. (\textbf{a}) The player communicates in natural language and optionally adds markers to the map to give absolute positions (Sec.~\ref{sec:a_interface}). (\textbf{b}) The LLM writes a plan following a structured output that controls the behavior of all units (Sec.~\ref{sec:b_plan}). (\textbf{c}) The plan assigns a target position and behavior tree to each unit when initialized or triggered by the objective checking (i.e., when a new step of the plan becomes active) (Sec.~\ref{sec:c_applying_plan}). (\textbf{d}) Each unit evaluates its behavior tree using local observation to determine an action (Sec.~\ref{sec:d_bts}). (\textbf{e}) After each step, the game checks the current plan's objectives  (Sec.~\ref{sec:e_check_plan}). If some are achieved, the plan is rolled forward, and the active steps are updated. This module also checks for global winning and losing conditions specific to each scenario. (\textbf{f}) At the start of the discussion, the LLM receives information about the types of each unit present (Sec.~\ref{sec:f_game_to_text}). When triggered by the player, the LLM is given a precomputed textual description of the map (see Sec.~\ref{sec:LLM_with_image} for a discussion of the geo-visual abilities of LLMs) and the current health and position of each unit. (\textbf{g}) After each game step, the units' positions are plotted on the map and displayed on the screen (Sec.~\ref{sec:g_visualization}).


\subsection{The multi-agent game}
In our top-down strategy game, the player commands an army of several hundred to thousands of units against an opposing army of comparable size. The player controls their army exclusively by interacting with HIVE, which generates a plan that determines each unit’s behavior at every step. The game is implemented in JaX~\cite{jax2018github}, a Python library for vectorization and just-in-time compilation. It allows fast parallelization of the units' behaviors, a key component for having an RTS game with many units. 

\subsubsection{Global objectives}
\label{sec:global_objectives}

There are two kinds of objectives: \emph{elimination objective} (i.e., eliminating a set of enemy units) and \emph{position objective} (i.e., reaching a target position). 

\subsubsection{The units}

Units are parameterized by their move speed, max health, attack damage, and attack range. We used three types: spearman, a slow, close-range unit with high health; archer, a long-range unit with low health; and cavalry, a fast, close-range unit with medium health. Those three types of units allow for rock-paper-scissors dynamics, as archers easily eliminate spearmen units using their longer range, cavalry units easily eliminate archer units by coming in contact faster than they get killed, and spearmen units easily win against cavalry units because of their higher health. Tab.~\ref{tab:units_type} in the appendix details their characteristics.

\subsubsection{The observation} 

Each unit knows the position and health of every unit inside its sight range (set to 15 m; sizes of the different environments are detailed in Section~\ref{sec:benchmark}). It also has access to a distance map (i.e., a matrix containing the distance to the unit's target position from anywhere on the map) to follow the shortest path to its target position. The map size in this paper ranges from 100~m to 300~m.

\subsubsection{The actions} 

At each step, each unit takes one action: do nothing, move to any position in sight reachable given the unit's speed (2D continuous action), or attack an enemy unit in attack range with a line of sight (discrete action). 

\subsubsection{The terrain} 

We use four types of terrain: 
\begin{itemize}
    \item \textbf{Normal}: the units can move and see through it;
    \item \textbf{Forest}: the units can move but not see through it;
    \item \textbf{Water}: the units can see over it but not move through it;
    \item \textbf{Building}: the units cannot move or see through it.
\end{itemize}
Those allow the creation of diverse maps and scenarios with strategic features such as choke points over bridges (normal terrain over water) or stealth opportunities by hiding in forests. We generated the maps with the INKARNATE platform~\cite{inkarnate}.

\subsubsection{The visualization}
\label{sec:g_visualization}
At each time step, the game plots the units' positions over the map using different shades of colors (blues for the player's units and reds for the enemy units) and shapes (square for the spearmen, circle for the archers, and triangle for the cavalry). The player does not receive information about each unit's health, as it would be difficult for them to assess the health of hundreds of units~\cite{sevcenko2023}.\\\\
In summary, the main game loop proceeds as follows: 
\begin{enumerate}[(a)]
    \item Apply the attack action of each attacking unit;
    \item Apply the move action of each moving unit while checking for collision with buildings or water terrain;
    \item Push units in collision with each other;
    \item Check for collision with a building or water terrain;
    \item Compute the distance between each unit, considering the broken line of sight over buildings and forest terrain.
\end{enumerate}

\subsection{Behavior Trees}

\label{sec:d_bts}
We use behavior trees (BTs)~\cite{colledanchise2018behavior} as a middle layer between the high-level instruction of the player, written down as a plan, and the low-level actions taken by the units. They offer a convenient way to control agents while being easily designed by hand. Alternative methods are discussed in Section~\ref{sec:discussion}.

BTs are composed of four different types of nodes: 
\begin{itemize}
    \item \textbf{Sequence node S}: stops at the first failed node;
    \item \textbf{Fallback node F}: stops at the first successful node;
    \item \textbf{Condition node C}: evaluates a condition given the unit observation and returns failure or success;
    \item \textbf{Action node A}: tries to perform an action given the unit observation and returns failure or success.
\end{itemize}

For this paper, we designed a list of condition and action nodes parametrized by qualifiers (see Sec.~\ref{sec:BT_grammar} in the appendix for the full Lark grammar). The available conditions are: 
\begin{itemize}
    \item Is there an ally or enemy of a given type in sight?
    \item Is there an ally or enemy of a given type in sight that is or could be in attack range in one, two, or three steps?
    \item Am I or could I be in attack range in one, two, or three steps of an ally or enemy of a given type in sight?
    \item Am I of a given type?
    \item Do I have health below a 75\%, 50\%, or 25\% threshold?
    \item Am I in a forest?
\end{itemize}
The available actions are:
\begin{itemize}
    \item Do nothing;
    \item Move toward the target position (following the shortest path with some added noise) up to a given threshold relative to the unit sight range and move speed;
    \item Move toward the closest or farthest or weakest or strongest or random ally or enemy unit in sight of a given type (without optimal path guarantee);
    \item Attack the closest or farthest or weakest or strongest or random enemy in attack range.
\end{itemize}

We handcrafted five behavior trees available to HIVE:
\begin{itemize}
    \item \textbf{Long-range attack}: move away from enemies, or attack an enemy if possible, or move toward the target position;
    \item \textbf{Close-range attack}: attack a random unit in attack range if possible or move toward the closest enemy if there is any; otherwise, move toward the target position;
    \item \textbf{Attack and move}: (1) attack a unit in attack range if possible or move toward the target position, and (2) only if the target is reached, move toward the closest enemy;
    \item \textbf{Move toward target position}: follow the shortest path to the target position with some noise;
    \item \textbf{Stand}: do nothing.
\end{itemize}
In all behavior trees concerning enemy units, the LLM can target any type of unit or subset of units. The exact syntax of all the behavior trees can be found in the appendix Sec.~\ref{sec:available_bts}.

\subsection{The language interface between the player and HIVE}
\label{sec:a_interface}
The player can only control their army through an interface with HIVE. Its front end is composed of an LLM that receives natural language prompts from the player and can reply to them. HIVE gives the LLM  task-specific instructions via a \textit{System} prompt, making it easy to use different LLMs by calling a different API.

Moving many units to different positions on a map can be conveniently and quickly realized through natural language commands. For example, asking the system to send them to cover all bridges can be faster than using a traditional mouse interface. However, there are situations where it can still be helpful to communicate a specific absolute map position to the model. Therefore, HIVE also includes the ability for the player to click on the map to add labeled markers that can then be referenced in the dialog with the model.

The outer loop unfolds as follows, HIVE: (1) initializes the game state; (2) waits for the user's prompt and sends it alongside the game state to the LLM; (3) waits for the LLM to provide a plan and parses it; and (4) executes the plan if it is valid. HIVE never asks the LLM to update the plan or try again. All the plans are saved, allowing us to re-execute them anytime for debugging or visualization using the same random seed, ensuring deterministic behavior.

\subsection{The plan}
\subsubsection{Writing a plan with our structured outputs}
\label{sec:b_plan}
HIVE translates the player's high-level commands into a plan that follows a structured output. We instruct the LLM (using a \textit{System} prompt) about the expected output format to write a plan with a simple example and a short list of mistakes to avoid. The player is not required to know the plan's format. One main advantage of HIVE is that it delegates the tedious part of writing the plan from the player to the LLM.

A plan is composed of several steps, and a step is composed of several components:
\begin{itemize}
    \item A numerical label to refer to it;
    \item A list of prerequisite steps that need to be achieved before this step (e.g., waiting for two groups to reach their assigned locations before ordering them to attack);
    \item An objective (similar to the global objectives) that is used to infer if the step is achieved, active, or inactive;
    \item A list of unit groups that dictate the behaviors of each unit concerned by the step.
\end{itemize}

A unit group comprises a list of unit IDs, a target position, and a behavior tree. The step is valid only if each unit is included in no more than one group. 

The actual instruction message used for Fig.~\ref{fig:teaser_figure} is detailed in the appendix Sec.~\ref{sec:appendix_instruction}. The plan written by the LLM is parsed, and an error is raised if it is not grammatically correct or if the units' or behavior trees' IDs are incorrect. 

\subsubsection{Applying the plan}
\label{sec:c_applying_plan}
When the plan is created or the active steps change, HIVE passes through each active step and assigns a behavior tree and target position to each unit. A unit keeps its previous assignment if it does not have a new one. If it never receives an assignment, it does nothing by default. 

This part constitutes HIVE's computational bottleneck since (1) each target position requires computing a breadth-first search on the map, and (2) reassigning the behavior trees and the steps' objective checking requires a new compilation of the JAX functions. However, most plans only require a handful of steps with no change in behavior trees between each transition, thus not affecting most game steps. 

\subsubsection{Checking the plan}
\label{sec:e_check_plan}
After each game step, HIVE checks the global and active step objectives to determine if one side is victorious, if the plan is fully unrolled, or if new steps become active. If the plan is fully unrolled, but no side is victorious, we stop the game and evaluate the plan as an early completion. 

\subsection{LLM game information}
\label{sec:f_game_to_text}

\subsubsection{Unit information}
The LLM gets a complete description of the unit types and the current composition of both sides. After each player's prompt, it receives the units' health and positions. An example is shown in the appendix Sec.~\ref{sec:appendix_game_info}.

\subsubsection{Textual description of the map}
During preliminary experiments, we explored visual map representations but found that current all-purpose LLMs are inefficient in understanding our top-down map and often cannot accurately position the units, terrain, or landmarks. Thus, we rely on a precomputed map description that gives the absolute position of the different terrain types (forest and rivers) and bridge positions. Sec.~\ref{sec:LLM_with_image} compares HIVE's abilities with the textual prompts and with different variations of inputting an image of the map. Fig.~\ref{fig:images_comparison_inputs_exploit_terrain} and \ref{fig:images_comparison_inputs_strategize_points} in the appendix show two examples of such descriptions for two maps used in the benchmark.

\section{Multi-agent control benchmark }
\label{sec:benchmark}

\begin{figure*}[t]
    \centering
    \includegraphics[width=\linewidth]{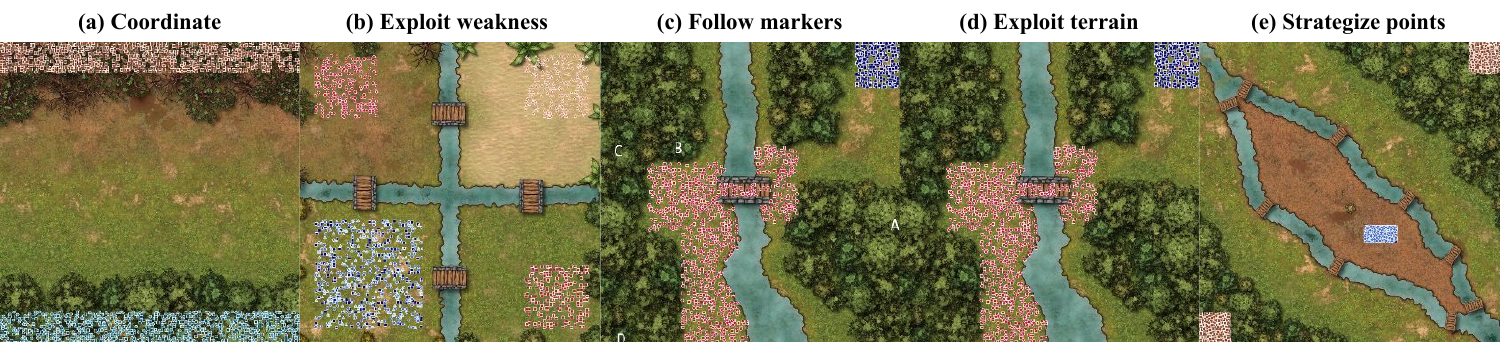}
    \caption{
    \textbf{The HIVE Benchmark Ability Tests.} (a) \textbf{Coordinate} where the player has to eliminate all the enemies using 1,000 units, (b) \textbf{Exploit weakness} where the player has to efficiently use the three types of units to eliminate the enemies, (c) \textbf{Follow markers} where the player has to bring at least one unit to the south, (d) \textbf{Exploit terrain} where the player has to bring at least one unit to the opposite corner of the map, and (e) \textbf{Strategize points} where the player has to prevent the enemies from reaching the center of their camp.}
    \label{fig:abilities_tests_maps}
\end{figure*}

    
\begin{figure*}[p]
    \centering
    \includegraphics[width=0.925\linewidth]{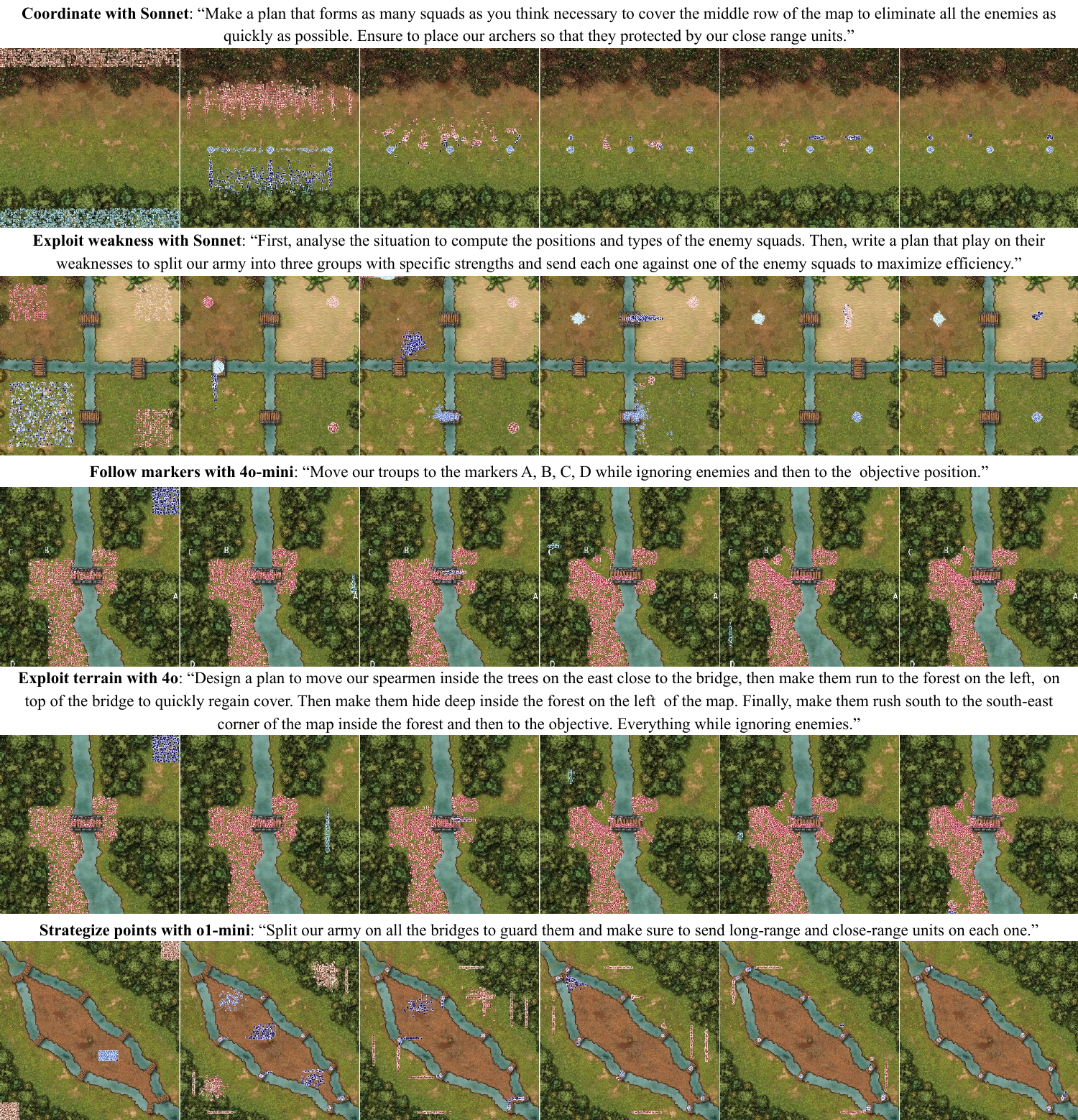}
    \caption{\textbf{Successful plans by HIVE using different LLMs for each ability test.} HIVE can translate high-level commands into successful plans, from coordinating thousands of units to exploiting the terrain, enemy weaknesses, or strategic points. Blue units are the player's units. Red units are the enemies.}
    \label{fig:main_snapshots}
\end{figure*}

\begin{figure*}[p]
    \centering
    \includegraphics[width=\linewidth]{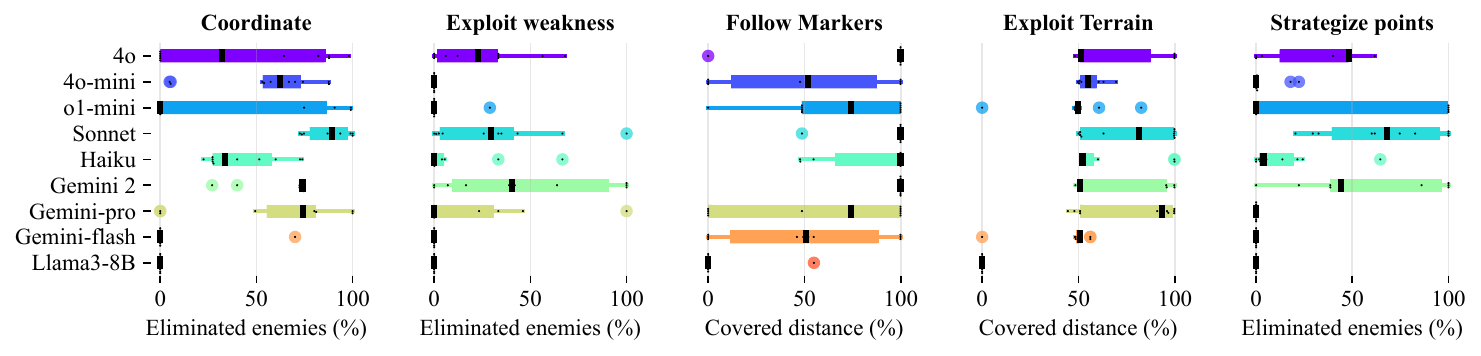}
    \caption{Ability evaluations of HIVE using nine LLMs with the same ten prompts for each ability test. Apart from Follow Markers, the different ability tests are solvable but challenging for all LLMs. \textbf{Sonnet} performs best, while a small LLM like Llama3-8B completely fails. }
    \label{fig:abilities_evaluation}
\end{figure*}

\subsection{The five ability tests}
\label{sec:benchmark_5_abilities}
We designed five ability tests on four scenarios (Fig.~\ref{fig:abilities_tests_maps}). All the scenarios use the same enemy units' behaviors: the spearmen and cavalry perform close-range attacks; the archers make long-range attacks while moving away from close combat. HIVE sends a pre-written prompt (along with state-specific information) to the LLM, parses the returned plan, and executes it (if it is valid) for a maximum number of game time steps, depending on the scenario. If, after this time, no side has won, the episode concludes in a tie. As all scenarios are designed to be winnable within the allocated time steps, a tie is better than a loss but worse than actually winning the scenario. If the plan is fulfilled without achieving the global objective, we call it an early completion. In all cases, we do not ask the LLM to propose a new plan.

\subsubsection{\textbf{Coordinate}}
The player controls 1,000~units (500~spearmen and 500~archers) and must eliminate the opposite army of 1,000 spearmen coming from the north in less than 300 time steps. This test thus asses the ability to coordinate many units with two types in a simple terrain. With the units only able to see 15~m ahead on the 150-meter-wide map, one challenge is to spread the units on the battlefield so as not to miss any enemies. Another challenge is to place the archers behind the spearmen to minimize casualties.

\subsubsection{\textbf{Exploit weakness}}
The player controls 750~units (250~spearmen, 250~archers, and 250~cavalry) and must eliminate the opposing army of equal composition in less than 500 time steps. The 100-meter-wide map is divided into four quadrants by vertical and horizontal rivers crossed by bridges. The opposite army is divided by types into three quadrants. 
This tests the ability to exploit the rock-paper-scissors dynamics to win while minimizing casualties. This demands that HIVE estimate the opposing army battalions' positions and unit types, dispatching the fitting counter type (HIVE is given the strengths and weaknesses of each unit type as in-context information). 

\subsubsection{\textbf{Follow markers \& Exploit terrain}} 
The player controls 300~spearmen starting on the northeast corner of the 200-meter-wide map and must bring at least one to an objective position in the southwest corner in less than 500 time steps. The units must cross a bridge guarded by 1,200~opposite units (600~spearmen and 600~archers). The challenge is efficiently using the trees' cover (the units cannot fight in the forest) to minimize casualties. We use this scenario for two ability tests: \textbf{Follow markers}, which tests the ability to follow marker positions given by the player, and \textbf{Exploit terrain}, which tests the ability to follow the player's textual description of the path to follow through the different terrains.

\subsubsection{\textbf{Strategize points}}
The player controls 700~units (350~spearmen and 350~archers) and must prevent the opposite army of 900~spearmen from reaching the center of the camp (symbolized by the campfire) for up to 500 time steps. The opposite units are divided in the northeast and southwest corners of the 300-meter-wide map and are designed to split and cross over each bridge. This tests the ability to divide the army to cover strategic points.

\subsection{The tested LLMs}
\label{sec:the_tested_LLMs}
We compare nine LLMs: eight state-of-the-art closed-source models from the OpenAI, Claude, and Gemini families, and one small open model as a baseline: 
\begin{itemize}
    \item \textbf{4o}: gpt-4o-2024-11-20~\cite{achiam2023gpt};
    \item \textbf{4o-mini}:gpt-4o-mini-2024-07-18~\cite{achiam2023gpt};
    \item \textbf{oi-mini}: o1-mini-2024-09-12~\cite{achiam2023gpt}:
    \item \textbf{Sonnet}: claude-3-5-sonnet-20241022~\cite{TheC3};
    \item \textbf{Haiku}: claude-3-5-haiku-20241022~\cite{TheC3};
    \item \textbf{Gemini 2}: models/gemini-2.0-flash-exp~\cite{team2023gemini};
    \item \textbf{Gemini-pro}: gemini-1.5-pro~\cite{team2023gemini};
    \item \textbf{Gemini-flash}: gemini-1.5-flash~\cite{team2023gemini};
    \item \textbf{Llama3 (8B)}: Llama3 (8B)~\cite{dubey2024llama}.
\end{itemize}

We performed an ablation to study the effect of the LLM's temperature on the plan's validity and quality (see Fig.~\ref{fig:images_temperature} in the appendix). We concluded that high temperatures increase the chance of returning an invalid plan and settled on a minimal temperature of 0 (except for \textbf{oi-mini}, which was a preview with a fixed temperature of 1).  

\subsection{Results}
\label{sec:result_5_abilities}

While experimenting, we found that the LLMs' proposed plans were susceptible to small variations in the player prompt, i.e.\ just one word being different. To consider the variability, we generated ten slightly different prompts for each of the five abilities. For example, to test the \textbf{Coordinate} ability, we query each LLM with ten variations of the prompt: \emph{“Make a plan that forms as many squads as you think necessary to cover the middle row of the map to eliminate all the enemies as quickly as possible. Ensure to place our archers so that they are protected by our close-range units.”}, resulting in ten plans (\textbf{Coordinate} row in Tab.~\ref{tab:ability_strict_sucesses}). The slight variations are the orders of saying \textit{“Design a plan,”} \textit{“Write down a plan,”} or \textit{“Make a plan,”} using \textit{“North/South/East/West”} or \textit{“Top/Down/Left/Right,”} or saying \textit{“archers units”} or \textit{“long-distance units.”} All 50 prompts are detailed in appendix Sec.~\ref{sec:prompts_appendix_5_ability_tests}. 

\begin{figure}[ht]
    \centering
    \includegraphics[width=\linewidth]{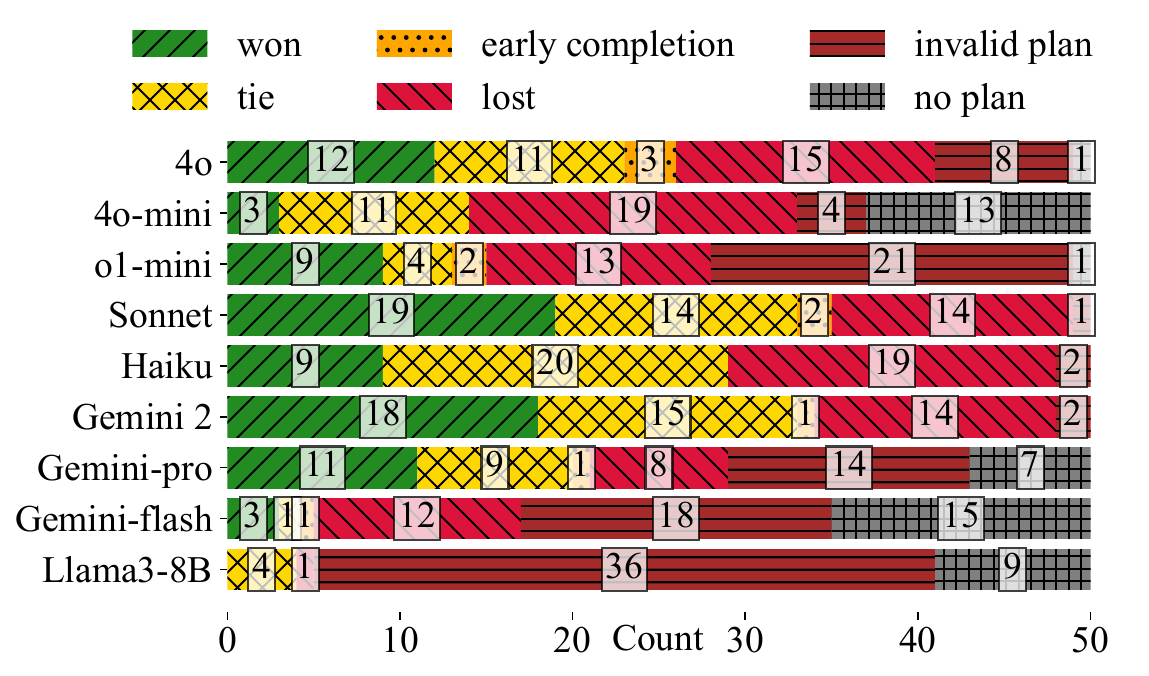}
    \caption{The summarized result of the ability tests of HIVE using each LLM.}
    \label{fig:plan_ending_bar_plot}
\end{figure}

Fig.~\ref{fig:main_snapshots} shows snapshots of a successful plan made by HIVE using different LLMs with the corresponding prompt from the player. The actual plans are detailed in the appendix Sec~\ref{sec:LLM_answer_and_plan}. Tab.~\ref{tab:ability_strict_sucesses} presents the strict successes of HIVE for the 10 prompt variations for each ability test. Fig.~\ref{fig:abilities_evaluation} presents a continuous evaluation of the performance of the plans made by HIVE. Using the percentage of enemies eliminated for the abilities (a), (b), and (e), and the distance to the objective position for (c) and (d). 
Fig.~\ref{fig:plan_ending_bar_plot} shows for each 50 prompts the result of the plan execution in terms of wins, losses, ties, achieving the plan objectives without achieving the winning objective, and HIVE returning no plan or an invalid plan. A video showing the executed plans is available at \url{hive.syrkis.com}.

HIVE solves all ability tests with \textbf{Sonnet} for at least one prompt variation, highlighting that it effectively possesses all the proposed abilities. It can effectively split the units on a battlefield, putting the long-range units in the rear to increase efficiency. Additionally, it can recognize clusters of enemy units with specific weaknesses and exploit them. Furthermore, HIVE can follow a strategic plan to exploit the cover of the trees as much as possible. Finally, it can also split the army to cover all the strategic points that the player asks. 

Concerning the LLMs' inference time (Fig.~\ref{fig:walltimes} in the appendix), excluding \textbf{o1} (which uses thinking) and \textbf{Llama3 (8B)} (which was run locally on an M3 chip), the models need several seconds to respond, e.g., \textbf{4o} with 6.4~s [3.3~s, 12.2~s] (median [first quartile, third quartile]), \textbf{Sonnet} with 12.6~s [9.9~s, 14.2~s], and \textbf{Gemini 2} with 3.9~s [3.0~s, 6.2~s].

The main takeaways from this evaluation are the following:
\begin{itemize}
    \item \textbf{Llama3 (8B)} cannot follow the plan structured output;
    \item As can be expected, \textbf{Follow markers} is the easiest ability test (\textbf{4o} and \textbf{Sonnet} solved it nine times out of ten and \textbf{Gemini 2} solved it every time);
    \item The other ability tests propose interesting challenges for LLMs as none of them got more than 50\% success;
    \item \textbf{Sonnet} is the only LLM to succeed in all five ability tests and only presents one invalid plan out of 50 answers;
    \item For the continuous evaluations, \textbf{Sonnet} is also either first (in \textbf{Coordinate} and \textbf{Strategize points}) or tied-first (in \textbf{Exploit weakness} and \textbf{Follow markers}); 
    \item \textbf{Haiku}, \textbf{4o-mini}, and \textbf{Gemini-flash} show, on average, worse performance than their full-version counterparts;
    \item The inference time is in the order of 1 to 10 s (see appendix Section~\ref{sec:appendix_inference_time});
    \item Even state-of-the-art LLMs are sensitive to the wording of the prompts, as slight changes in the prompt result in drastic changes in the plan and the execution.
\end{itemize}

\begin{table*}[ht]
\caption{Ability tests of HIVE with different LLMs showing the success of achieving the global objective of each scenario.}
\centering
\label{tab:ability_strict_sucesses}
\begin{tabular}{@{}rccccccccc@{}}
\toprule
\multicolumn{1}{c}{}       & \textbf{4o}                        & \textbf{\begin{tabular}[c]{@{}c@{}}4o\\ mini\end{tabular}} & \textbf{\begin{tabular}[c]{@{}c@{}}o1\\ mini\end{tabular}} & \textbf{Sonnet}                    & \textbf{Haiku}                    & \textbf{Gemini 2}                  & \textbf{\begin{tabular}[c]{@{}c@{}}Gemini\\ pro\end{tabular}} & \textbf{\begin{tabular}[c]{@{}c@{}}Gemini\\ flash\end{tabular}} & \textbf{\begin{tabular}[c]{@{}c@{}}Llama\\ 3 (8B)\end{tabular}} \\ \midrule
\textbf{Coordinate}        & 0\tiny/10           & 0\tiny/10                                   & 0\tiny/10                                   & 2\tiny/10           & 0\tiny/10          & 0\tiny/10              & 2\tiny/10                & 0\tiny/10                  & 0\tiny/10                               \\
\textbf{Exploit weakness}  & 0\tiny/10           & 0\tiny/10                                   & 0\tiny/10                                   & 1\tiny/10           & 0\tiny/10          & 3\tiny/10              & 1\tiny/10                & 0\tiny/10                  & 0\tiny/10                               \\
\textbf{Follow markers}    & 9\tiny/10           & 3\tiny/10                                   & 5\tiny/10                                   & 9\tiny/10           & 7\tiny/10          & 10\tiny/10             & 5\tiny/10                & 3\tiny/10                  & 0\tiny/10                               \\
\textbf{Exploit terrain}   & 3\tiny/10           & 0\tiny/10                                   & 0\tiny/10                                   & 4\tiny/10           & 2\tiny/10          & 2\tiny/10              & 3\tiny/10                & 0\tiny/10                  & 0\tiny/10                               \\
\textbf{Strategize points} & 0\tiny/10           & 0\tiny/10                                   & 4\tiny/10                                   & 3\tiny/10           & 0\tiny/10          & 3\tiny/10              & 0\tiny/10                & 0\tiny/10                  & 0\tiny/10                               \\
\textbf{Total}             & \textbf{12\tiny/50} & \textbf{3\tiny/50}                          & \textbf{9\tiny/50}                          & \textbf{19\tiny/50} & \textbf{9\tiny/50} & \textbf{18\tiny/50}    & \textbf{11\tiny/50}      & \textbf{3\tiny/50}         & \textbf{0\tiny/50}                      \\ \bottomrule
\end{tabular}
\end{table*}

\section{HIVE ablation studies}

\subsection{How does HIVE scale up with the number of units?}
\label{sec:llm_scaling}

\begin{figure}[ht]
    \centering
    \includegraphics[width=\linewidth]{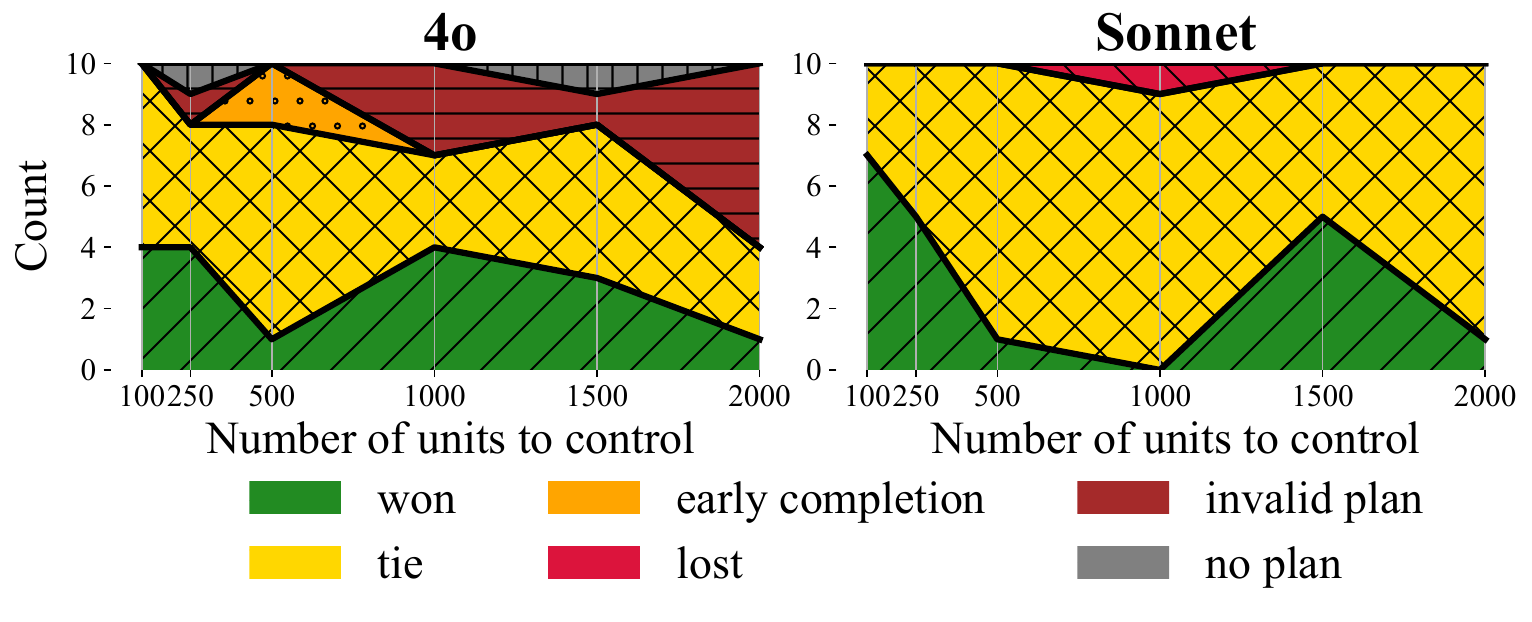}
    \caption{\textbf{Unit Scaling.} Comparison of the results of the HIVE plan for the Coordinate test by varying the number of units in the game.}
    \label{fig:llm_scaling}
\end{figure}

To see how HIVE scales with the number of units, we performed another evaluation on the \textbf{Coordinate} test, varying the number of units from 200 to 4,000. We stopped at 4,000 due to the current hardware limitation of the machine running the experiment. As the game's main loop needs to be compiled, the size of the distance matrix between each unit grows quadratically with the number of units. For this study, we used \textbf{Sonnet}, which showed the best performance, and \textbf{4o} to see if the results generalize to a different model.

Fig.~\ref{fig:llm_scaling} shows the result of the plan made by HIVE. There is no significant conclusion regarding the success rate, as there is too much noise due to the inherent variance of the LLMs. \textbf{Sonnet}'s large number of ties is due to its plans not successfully covering the frontline, missing some enemy units, with units of both sides waiting at different locations. Still, \textbf{4o}'s number of invalid and empty plans increases as the number of agents increases. This limitation is not observed for \textbf{Sonnet}, again highlighting its superior performance.


\subsection{Can HIVE win alone?}
\label{sec:llm_alone}
An interesting question is how well HIVE performs without a human's help. To answer this question, we use ten prompt variations: \textit{"First, analyze the situation to find a good strategy to win this mission, then write down the corresponding plan."} (the variations are detailed in the appendix Sec.~\ref{sec:HIVE_alone_appendix}) on each of the four scenarios for \textbf{Sonnet} and \textbf{4o}. HIVE always describes the scenario so the LLM knows the global objective.

\begin{figure}[ht]
    \centering
    \includegraphics[width=\linewidth]{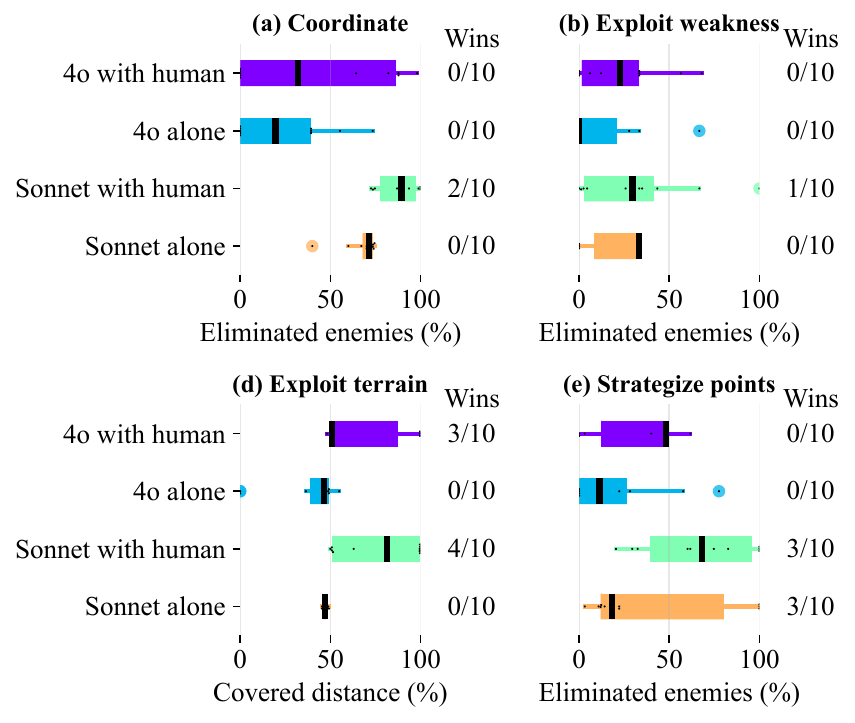}
    \caption{\textbf{Human-Machine Collaboration.} Without the player's help, HIVE's performances and win rates decrease for both models in all ability tests.}
    \label{fig:llm_alone}
\end{figure}

Fig.~\ref{fig:llm_alone} compares HIVE with and without the player's help. For \textbf{Coordinate} and \textbf{Exploit terrain}, both models no longer win without human help. For \textbf{Strategize Points}, \textbf{Sonnet} conserves its three wins, but both models have lower median performances. These results show that HIVE is better with human help, confirming the benefits of a hybrid intelligence approach. Additionally, when comparing the median performance, both \textbf{Coordinate} and \textbf{Exploit Weakness} seem to be easier ability tests for HIVE alone, as they only require sending units toward the enemies. In comparison, \textbf{Exploit terrain} and \textbf{Strategic points} require exploiting the map features to win and depend on more long-term planning.

\subsection{Does HIVE need the textual description of the map?}
\label{sec:LLM_with_image}

\begin{figure}[ht]
    \centering
    \includegraphics[width=\linewidth]{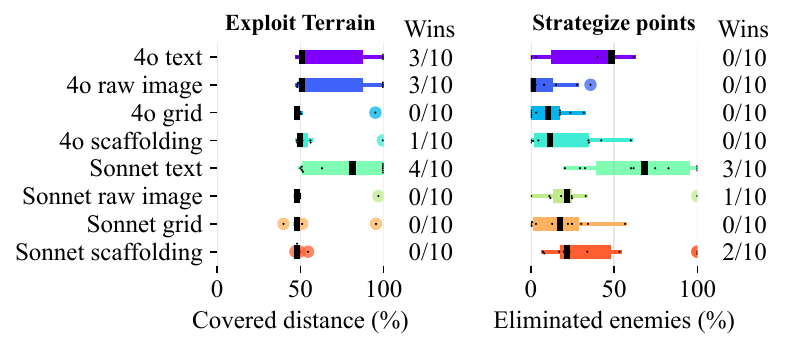}
    \caption{\textbf{Maps description.} LLMs' performance in using maps drops when switching from textual descriptions to images (raw, with a grid, or scaffolding). }
    \label{fig:images_comparison}
\end{figure}


We evaluate HIVE's vision abilities by replacing the textual description of the map with the image. As it has to use precise target positions, we also compared two ways to improve the image-to-coordinate: a grid with an x and y axis and a scaffolding~\cite{lei2024scaffolding}. We only tested the vision on two ability tests that required it: \textbf{Exploit terrain} and \textbf{Strategize points}. The four kinds of inputs are shown in the appendix Section~\ref{sec:appendix_map_description}.

Without the textual description, HIVE's performance decreases (Fig.~\ref{fig:images_comparison}). For \textbf{Exploit terrain}, \textbf{4o} can exploit the raw image, while \textbf{Sonnet} fails to do so. For \textbf{Strategize points}, \textbf{Sonnet} won once with the raw image and twice with the scaffolding,  with lower median performances than the three victories with the textual description.

\section{Discussion}
\label{sec:discussion}

None of the closed-source LLMs were fine-tuned to the plan's structured output or our game. Still, they returned valid and successful plans, demonstrating that HIVE can be used with an all-purpose LLM. As \textbf{Llama3-(8B)} mostly returns invalid plans, closed-source models currently achieve the best performance on the proposed benchmark.

An exciting future prospect is to take a smaller model and fine-tune it, allowing HIVE to run locally. Together with dedicated hardware, this could enable the system to be fast enough to run continuously alongside the game to (1) give real-time feedback to the player, (2) adjust the plan as the game unrolls, or (3) even run beyond real-time for fast self-play. While becoming increasingly efficient at coding in widely used programming languages, LLMs still lag behind for less common languages and specific structured outputs, mainly due to the restricted available training data~\cite{joel2024survey}. Future research is still needed to propose efficient fine-tuning solutions.

The poor results of HIVE on the \textbf{Exploit weakness} task can be explained by the LLMs' inability to extract the actual positions to target from the units' raw positions given as context. This corroborates that LLMs struggle to process mathematical data~\cite{wang2024d, oh2025}. One direction is to improve LLMs' skills for numerical analysis. Another is to design (or learn) a feature selector that summarizes the unit's information.

The inference time of the LLMs is between four and 12 seconds. This delay is problematic for our game, where the average plan execution lasts between 10 and 20 seconds. There, we decided to only perform one plan inference per game. However, this inference time would be less of an issue in games where the average game lasts more than a few minutes. Additionally, LLM performances are improving rapidly. For example, \textbf{Gemini 2}, which was designed for low latency, is twice as fast as \textbf{Sonnet} and only has one less victory. Another direction to decrease the inference time is to use summarization tools~\cite{widyassari2022review} to shrink the context size. 

Following the conclusion of our temperature ablation, we decided to set the temperature to 0. A recent study of the effect of the temperature for code generation with ChatGPT~\cite{ouyang2025a} highlights that a temperature of 0 reduces the non-determinism but does not guarantee strict determinism. Non-determinism must be considered for tasks requiring long-term planning, such as ours. One direction for our hybrid framework is allowing the player to review the plan before execution to correct any possible mistakes. 

We used a structured output format to define the plan that assigns predefined behavior trees to each unit. One alternative is to generate code directly, as it has been done for games like Minecraft~\cite{wang2023voyager}. We hypothesized that focusing on a simple, structured format for a plan using behavior trees would allow for easy interpretation of both the high-level plan and the low-level behavior of each unit. In contrast, understanding the code generated by the LLM could be difficult for non-programmers. In addition, having a structured format allows us to constrain the available strategies and behaviors more easily.

Incorporating findings of recent research focusing on improving LLMs' interaction in a GUI environment~\cite{liu2025pc} and geospatial reasoning~\cite{zhang2024geogpt} could further improve HIVE's ability to handle multi-modal inputs from the player and the game interface. A recent benchmark for using LLM based on vision for simple adversarial games (e.g., Tic-Tac-Toe, Reversi, Gomoku, and Chess) shows that current models struggle to perform well~\cite{wanglarge}, highlighting the need for further research.

Our benchmark results show limitations of current LLMs; nonetheless, it also discriminates between the models, as Claude Sonnet 3.5 and Gemini 2.0 are less impacted than their GPT-4o. Highlighting the rapid progress in the field, just before the publication of this article, we ran a preliminary comparison on the just-released Sonnet 3.7. This model passes the 50\% success mark with 26 victories out of 50, emphasizing the need for the community to build benchmarks such as ours to keep highlighting their progress and limitations. 

\section{Conclusion}
This paper presents a new challenge for LLMs as human assistants to control many units in a strategy game. We proposed a new framework, HIVE, to allow a player to give high-level commands that an LLM translates into a long-term plan that controls the behavior of each unit. We showed that generalist LLMs such as Claude Sonnet 3.5, Gemini 2, and GPT-4o can handle such tasks but are still sensitive to slight changes in the player's prompts. Complementary experiments showed that HIVE requires human help to get the best performance and that generalist LLMs' visual capacity to use an out-of-distribution map for terrain and landmark locations still needs improvement. This work opens many interesting avenues for enhancing LLMs' capacities to collaborate with humans, such as better map-reading abilities and long-term planning, as well as reduced sensitivity to prompt variations.

\section*{Acknowledgments}
Funded by the armasuisse S+T project F00-007.

\bibliographystyle{IEEEtran}
\bibliography{biblio}

\clearpage
\newpage
\appendix

\subsection{Inference time analyses}
\label{sec:appendix_inference_time}

%
%

\begin{figure}[ht]
    \centering
    \includegraphics[width=\linewidth]{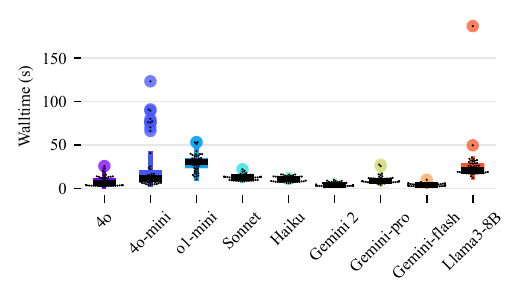}
    \caption{\textbf{Wall-times of the 50 LLM queries of the main evaluations.} \textbf{Llama3-8B} is run on an M3 chip. HIVE queries the others via their API.}
    \label{fig:walltimes}
\end{figure}

\subsection{Temperature ablation}
\label{sec:appendix_temperature}
\begin{figure}[ht]
    \centering
    \includegraphics[width=\linewidth]{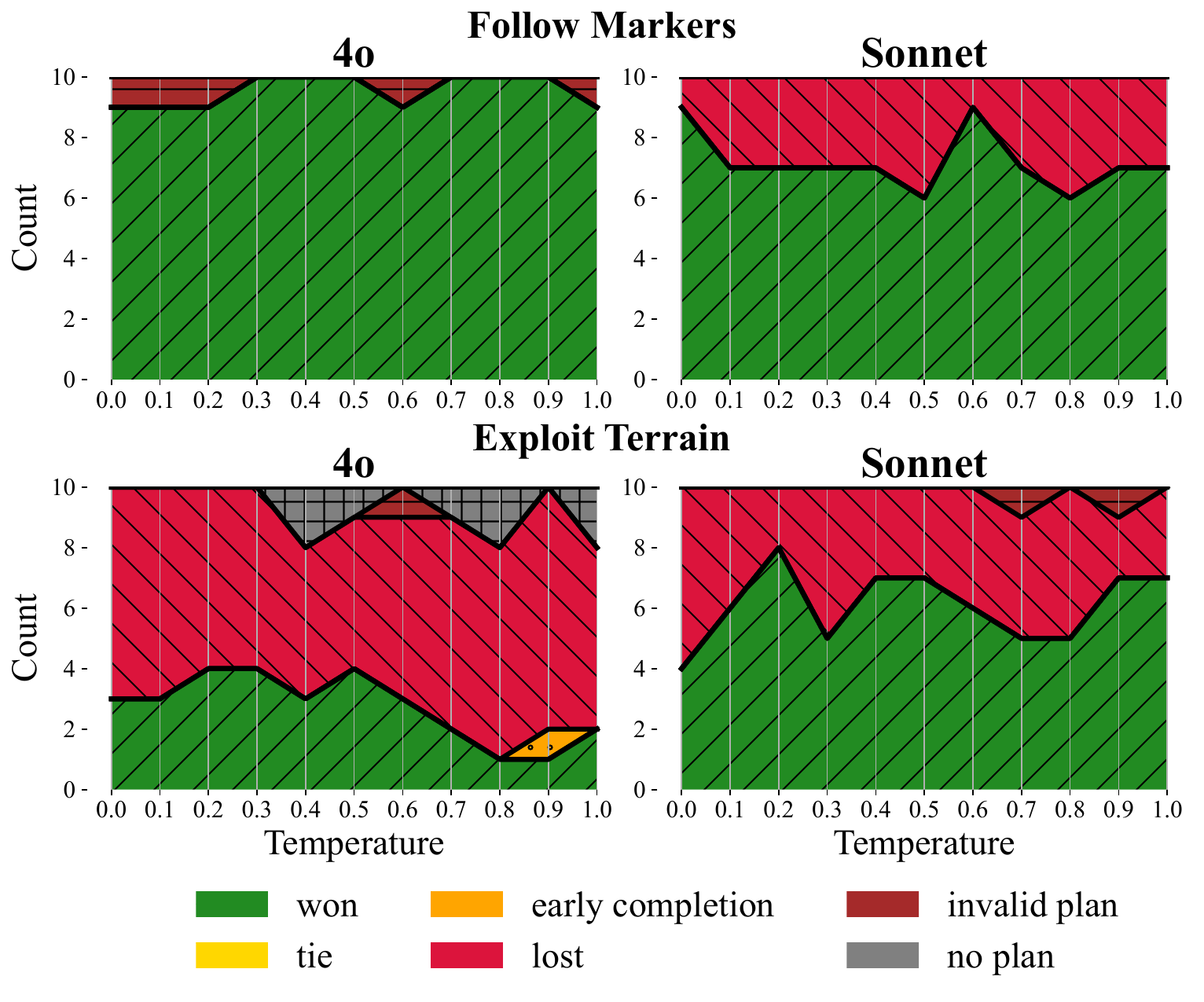}
    \caption{LLM temperature effect on the plan execution for \textbf{4o} and \textbf{Sonnet} on the \textbf{Follow markers} and \textbf{Exploit terrain} ability tests.}
    \label{fig:images_temperature}
\end{figure}

We performed an ablation study of the effect of the LLM temperature (Fig.~\ref{fig:images_temperature}).
Increasing the temperature doesn't significantly affect the plans' quality for the simpler ability test \textbf{Follow markers}. For the more complex ability test, \textbf{Exploit terrain}, \textbf{4o} starts returning no plan 2 out of 10 queries with a temperature above 0.3 and has a slight decrease of victorious plans with temperatures above 0.6, and \textbf{Sonnet} starts returning invalid plans 1 out of 10 queries for temperatures above 0.6.
 
\subsection{Map description inputs}
\label{sec:appendix_map_description}
\begin{figure}[ht]
    \centering
    \includegraphics[width=\linewidth]{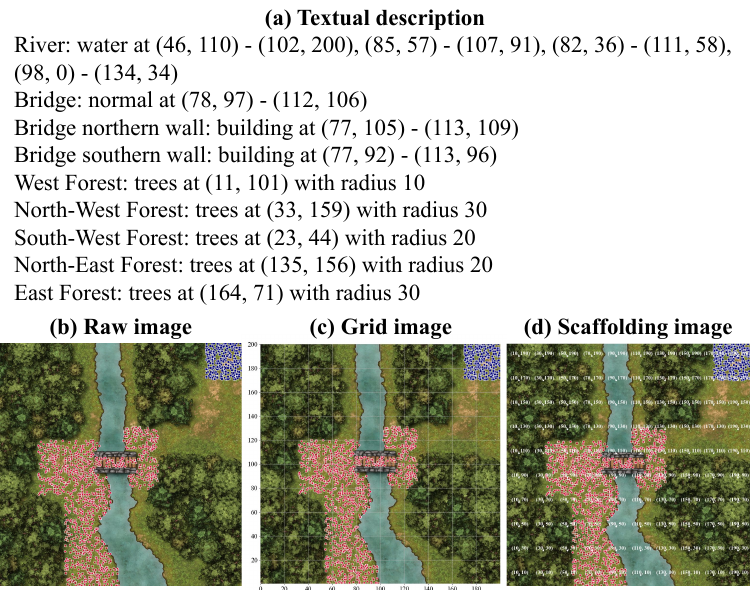}
    \caption{\textbf{Inputs describing the map for the Exploit terrain ability test.}}
    \label{fig:images_comparison_inputs_exploit_terrain}
\end{figure}

\begin{figure}[ht]
    \centering
    \includegraphics[width=\linewidth]{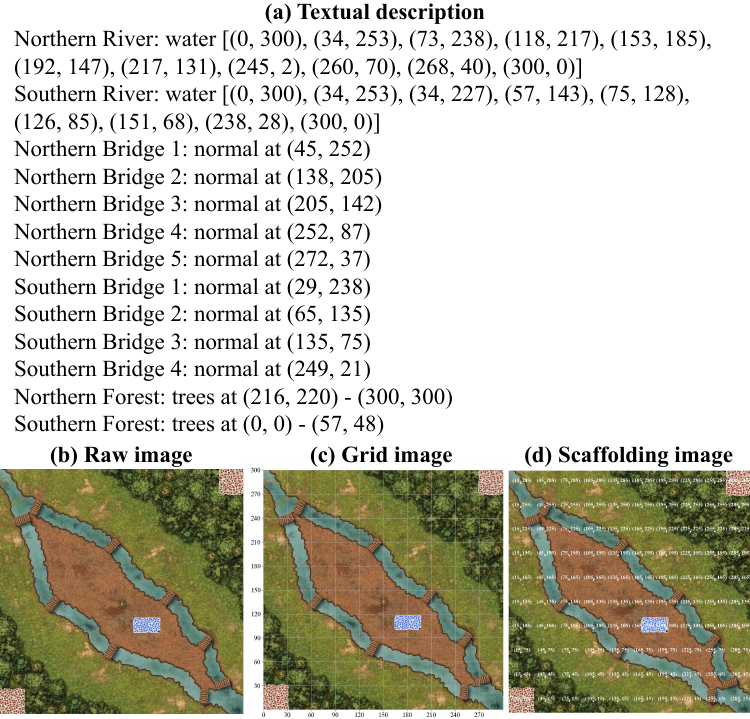}
    \caption{\textbf{Inputs describing the map for the Strategize points ability test.}}
    \label{fig:images_comparison_inputs_strategize_points}
\end{figure}

The section presents the four inputs used for the "Does HIVE need the textual description of the map?" ablation. Fig.~\ref{fig:images_comparison_inputs_exploit_terrain} for the \textbf{Exploit terrain} ability test and Fig.~\ref{fig:images_comparison_inputs_strategize_points} for the \textbf{Strategize points} ability test.


\subsection{Units characteristics}
\begin{table}[ht]
\caption{Characteristics of the three types of units used in this paper.}
\label{tab:units_type}
\centering
\begin{tabular}{@{}rccc@{}}
\toprule
\multicolumn{1}{l}{}        & Spearman & Archer & Cavalry \\ \midrule
Visual shape                       & square   & circle &  triangle    \\ 
Speed (unit distance/step)  & 1        & 2      & 6       \\
Max health                  & 24       & 2      & 12      \\
Attack damage               & 1        & 3      & 1       \\
Attack range                & 1        & 15     & 1       \\
\bottomrule
\end{tabular}
\end{table}

\subsection{LLM instruction and game for the example of the teaser figure}

\subsubsection{Instruction}
\label{sec:appendix_instruction}
\begin{Verbatim}[breaklines=true]
    # Map Instruction
    
You are a game assistant that helps the player in a strategy video game. 
Your task is to discuss with the player to come up with a plan to win the game's scenario (which will be provided later). Work with the player by giving feedback about their propositions, asking questions to clarify, obtain more details, or decide between different propositions. The player can also ask you questions.
    

You will be given a textual description of each pertinent element of the map using their name, terrain type, and bounding boxes (bottom-left corner - top-right corner).
In the game, there are four types of terrain:
- Normal: units can cross and see through (by default, the whole map is normal).
- Buildings: units cannot cross or see through buildings.
- Water: units cannot cross over water but can see over it.
- Trees: units cannot see through trees but can move over them. In particular, once a unit is inside a tree area, it cannot see any other unit.
    
Also, for simplicity, bridges that allow crossing water terrain will be specified. They correspond to normal terrain.
You can assume that any part of the map that is not included in any of the pertinent elements has a normal type of terrain.
A common convention is that East = Right, North = Top, West = Left, and South = Bottom. So the point (0, 0) is the bottom-left corner of the map. 
The x-axis increases from West to East = from Left to Right, and the y-axis increases from South to North = from Bottom to Top.
    
Format: 
[Name]: [terrain type] at [circle or square coordinate], [circle or square coordinate], ..., [circle or square coordinate]
where [circle] = [(x, y) coordinates of center] with radius [R]
where [square coordinate] = [(x, y) coordinates of the bottom-left corner] - [(x, y) coordinates of the top-right corner]

### For example 
East Forest: trees at (12, 63) with radius 20
North River: water at (0, 85) - (100, 90)
North River's Bridge: normal at (45, 85) - (55, 90)
West Forest: trees at (0, 23) with radius 6, (5, 33) with radius 7
    

    
## Markers
    
Through the discussion, the player can define markers on the map, which will be provided to you using the following format:
Markers:
A at (17, 5)
B at (6, 32)
C at (25, 28)

# Information About the Units
    

Description of the unit types in the game:
spearmen: Health=24; Sight range=15; Attack range=1; Moving speed=2; Attack damage=1; Attack cooldown=1
archer: Health=2; Sight range=15; Attack range=15; Moving speed=4; Attack damage=3; Attack cooldown=1
cavalry: Health=12; Sight range=15; Attack range=1; Moving speed=12; Attack damage=1; Attack cooldown=1
Spearmen are strong against Cavalry because they have more health.
    Cavalry are strong against Archers because they can quickly engage in close combat where Archers are weak.
    Archers are strong against Spearmen because they can attack from a longer distance.

    
Here is the list of unit IDs (in the form of a Python slice a:b with a included and b not included) for each team and the type of units they are composed of:

Descriptions of each team's composition:
Allies:
	spearmen: [0:350]
	archer: [350:700]
Enemies:
	spearmen: [0:900]

    
You will be provided with a list of all the units' current health and position.
Both teams are composed of many units. You should start by analyzing the positions of the units still alive to compose groups and compute their average position so that you can form the plan and send units to the appropriate positions.


# How You Should Handle the Player's Prompt

The user will ask you to write one plan. If you already have a shared conversation and the user asks you to modify or add steps to the plan, take care to build upon the already selected plan.

# Syntax for a Detailed Plan

A detailed plan is a set of steps to achieve in a given order until all the steps are completed.
You must provide the steps of the plan between the two keywords "BEGIN PLAN" and "END PLAN". 
As you only propose one plan, you should use these two keywords once.

One step comprises:
- A numeral ID
- A list of prerequisite steps that need to be completed before the step is rolled out
- An objective
A succession (at least one) of groups of units:
- The unit IDs of the group
- Target position on the map for the units to go to if there are no enemies in sight
- A behavior for the units if there are enemies in sight

# Syntax for a Detailed Plan (continued)

The list of prerequisites corresponds to the list of steps' IDs that must be completed before trying to achieve the objective.

There are two kinds of objectives:
- **Elimination objective:** where the objective is completed when all the targets are eliminated. In that case, you must provide a list of enemies' IDs or the keyword "all" if all enemies are targets.
- **Position objective:** where the objective is completed when all the concerned units are close to their target position.

Position objectives are a good way to move your units, but if the end objective is to eliminate enemies, it can be more straightforward to directly set an elimination objective and use the target position to move the units.
The concerned units are either the keyword "all" (if all the allies' units are concerned) or a list of integers corresponding to the unit IDs.

The behavior corresponds to a low-level and local behavior that the units will follow.
Here is the list of available behaviors:
- **attack_in_close_range:** the unit attacks the enemies in close range if there are enemies or moves toward the target position if there are no enemies.
- **attack_and_move:** the unit attacks the enemies without moving if there are enemies or moves toward the target position if there are no enemies.
- **attack_in_long_range:** the unit attacks the enemies in long range if there are enemies or moves toward the target position if there are no enemies.
- **follow_map:** ignore the enemies and simply move straight to the target objective. Only use it when the player asks you to ignore the enemies.

Apart from standing still, all these behaviors will only be active if an enemy is in sight. Otherwise, they will move to a target position if you set one in the plan or stand still if no target position is set.
Remember that units collide and push each other, so ignoring the enemies may not be the fastest way to reach a target position if there are enemies on the way who could block you.

The syntax format of a step is the following:
Step ID: (where you replace ID with the integer ID of the step)
prerequisites: [s_1, s_2, ..., s_n] (where the s_i correspond to the prerequisites' steps IDs. Note that the list can be empty. In that case, simply write [])
objective: position [or] elimination UNIT_LIST
(At least one list of units and their assigned behavior and target position, but there can be as many groups as there are units, as one unit can belong to two groups:)
UNIT_LIST:
- target position: (x, y) (The integer x and y coordinate on the map.)
- behavior: behavior_name target_1 target_2 ... target_n (where behavior_name is an available behavior and target_1 to target_n are the targeted unit types or just the keyword "any" if the behavior targets any unit_types)

A UNIT_LIST is the list of unit IDs, and it has the following format:
- Either it's the word "all" (without quotes), which means that all the units of the corresponding team are concerned.
- Or it's a list of IDs in the format "[X1, X2, ..., Xn]" where Xi can either be an integer or a slice "a:b" (with a and b as integers, b>a, a included and b not included like in Python's range function). If a is not specified (i.e., ":b") it is considered to be 0, and if b is not specified (i.e., "a:") it is considered to be the total number of agents in the considered team.

### IMPORTANT:
- All positions (x, y) must be integers. If you want to give float positions, convert them first into integers.
- Do not add comments to the plan specification, as it interferes with the parser. If you want to give comments, give them outside the "BEGIN PLAN" and "END PLAN".
- A unit can only belong to one group of units for the same step.

## Example of a Valid Detailed Plan:

BEGIN PLAN
Step 0:
prerequisites: []
objective: position
units: all
- target position: (24, 14)
- behavior: attack_and_move any

Step 1:
prerequisites: [0]
objective: position
units: [0,1,2,10:]
- target position: (24, 16)
- behavior: attack_and_move any
units: [10:15,30:]
- target position: (24, 14)
- behavior: attack_and_move any

Step 2:
prerequisites: [1]
objective: elimination [:15]
units: [:30]
- target position: (24, 24)
- behavior: attack_in_close_range archer spearmen
units: [30:60]
- target position: (24, 24)
- behavior: attack_in_long_range spearmen
END PLAN

## List of Planning Mistakes You Should Avoid:

Avoid creating a series of position objectives with different groups of units waiting for each other (through a chain of prerequisites). Instead, regroup those positions into one step so that all units can move simultaneously.

Ensure all units have assigned behaviors at every step. If two or more steps can be active simultaneously (because they share the same prerequisites), each unit must belong to at least one group.

For elimination objectives, ensure the target position is close enough to the targeted enemy units for effective engagement.

Elimination objectives already have a default target position, so avoid unnecessary position objectives before them. Instead, set the target position of the elimination objective to the same location.

Ensure the units’ IDs are enclosed in square brackets [ and ].

Use the correct type name for long-range units as "archer" (not "archers").

Verify the name of the unit behaviors (e.g., "attack_in_close_range") and ensure all units' target positions are integers.

Be sure to adapt this format and guidance as needed when responding to the player's requests.
In this scenario, you control the Allies and must prevent the Enemies from invading the center of your camp in (150, 134). 
    You must elaborate a defensive plan that eliminates all the Enemies while preventing any of them from reaching the center of your camp.
    If one Enemy unit reaches the center of your camp, all your units are eliminated, or if your plan is completed while there are still standing Enemy units, you lose the game.
    Your intel reports inform you that the enemy will converge toward the center of your camp and attack your units in sight at close range.
\end{Verbatim}

\subsubsection{The game info}
\label{sec:appendix_game_info}
\begin{Verbatim}[breaklines=true, breakautoindent=true]
Health and positions of all the units of each team (∅ means that the unit is dead).
Allies:
Health: [24, 24, 24, 24, 24, 24, 24, 24, 24, 24, 24, 24, 24, 24, 24, 24, 24, 24, 24, 24, 24, 24, 24, 24, 24, 24, 24, 24, 24, 24, 24, 24, 24, 24, 24, 24, 24, 24, 24, 24, 24, 24, 24, 24, 24, 24, 24, 24, 24, 24, 24, 24, 24, 24, 24, 24, 24, 24, 24, 24, 24, 24, 24, 24, 24, 24, 24, 24, 24, 24, 24, 24, 24, 24, 24, 24, 24, 24, 24, 24, 24, 24, 24, 24, 24, 24, 24, 24, 24, 24, 24, 24, 24, 24, 24, 24, 24, 24, 24, 24, 24, 24, 24, 24, 24, 24, 24, 24, 24, 24, 24, 24, 24, 24, 24, 24, 24, 24, 24, 24, 24, 24, 24, 24, 24, 24, 24, 24, 24, 24, 24, 24, 24, 24, 24, 24, 24, 24, 24, 24, 24, 24, 24, 24, 24, 24, 24, 24, 24, 24, 24, 24, 24, 24, 24, 24, 24, 24, 24, 24, 24, 24, 24, 24, 24, 24, 24, 24, 24, 24, 24, 24, 24, 24, 24, 24, 24, 24, 24, 24, 24, 24, 24, 24, 24, 24, 24, 24, 24, 24, 24, 24, 24, 24, 24, 24, 24, 24, 24, 24, 24, 24, 24, 24, 24, 24, 24, 24, 24, 24, 24, 24, 24, 24, 24, 24, 24, 24, 24, 24, 24, 24, 24, 24, 24, 24, 24, 24, 24, 24, 24, 24, 24, 24, 24, 24, 24, 24, 24, 24, 24, 24, 24, 24, 24, 24, 24, 24, 24, 24, 24, 24, 24, 24, 24, 24, 24, 24, 24, 24, 24, 24, 24, 24, 24, 24, 24, 24, 24, 24, 24, 24, 24, 24, 24, 24, 24, 24, 24, 24, 24, 24, 24, 24, 24, 24, 24, 24, 24, 24, 24, 24, 24, 24, 24, 24, 24, 24, 24, 24, 24, 24, 24, 24, 24, 24, 24, 24, 24, 24, 24, 24, 24, 24, 24, 24, 24, 24, 24, 24, 24, 24, 24, 24, 24, 24, 24, 24, 24, 24, 24, 24, 24, 24, 24, 24, 24, 24, 24, 24, 24, 24, 24, 24, 24, 24, 24, 24, 24, 24, 2, 2, 2, 2, 2, 2, 2, 2, 2, 2, 2, 2, 2, 2, 2, 2, 2, 2, 2, 2, 2, 2, 2, 2, 2, 2, 2, 2, 2, 2, 2, 2, 2, 2, 2, 2, 2, 2, 2, 2, 2, 2, 2, 2, 2, 2, 2, 2, 2, 2, 2, 2, 2, 2, 2, 2, 2, 2, 2, 2, 2, 2, 2, 2, 2, 2, 2, 2, 2, 2, 2, 2, 2, 2, 2, 2, 2, 2, 2, 2, 2, 2, 2, 2, 2, 2, 2, 2, 2, 2, 2, 2, 2, 2, 2, 2, 2, 2, 2, 2, 2, 2, 2, 2, 2, 2, 2, 2, 2, 2, 2, 2, 2, 2, 2, 2, 2, 2, 2, 2, 2, 2, 2, 2, 2, 2, 2, 2, 2, 2, 2, 2, 2, 2, 2, 2, 2, 2, 2, 2, 2, 2, 2, 2, 2, 2, 2, 2, 2, 2, 2, 2, 2, 2, 2, 2, 2, 2, 2, 2, 2, 2, 2, 2, 2, 2, 2, 2, 2, 2, 2, 2, 2, 2, 2, 2, 2, 2, 2, 2, 2, 2, 2, 2, 2, 2, 2, 2, 2, 2, 2, 2, 2, 2, 2, 2, 2, 2, 2, 2, 2, 2, 2, 2, 2, 2, 2, 2, 2, 2, 2, 2, 2, 2, 2, 2, 2, 2, 2, 2, 2, 2, 
2, 2, 2, 2, 2, 2, 2, 2, 2, 2, 2, 2, 2, 2, 2, 2, 2, 2, 2, 2, 2, 2, 2, 2, 2, 2, 2, 2, 2, 2, 2, 2, 2, 2, 2, 2, 2, 2, 2, 2, 2, 2, 2, 2, 2, 2, 2, 2, 2, 2, 2, 2, 2, 2, 2, 2, 2, 2, 2, 2, 2, 2, 2, 2, 2, 2, 2, 2, 2, 2, 2, 2, 2, 2, 2, 2, 2, 2, 2, 2, 2, 2, 2, 2, 2, 2, 2, 2, 2, 2, 2, 2, 2, 2, 2, 2, 2, 2, 2, 2, 2, 2, 2, 2, 2, 2, 2, 2, 2, 2, 2, 2, 2, 2, 2, 2, 2, 2, 2, 2, 2, 2, 2, 2, 2, 2, 2, 2]
X positions: [180, 181, 189, 195, 170, 169, 186, 176, 166, 172, 194, 166, 191, 165, 190, 165, 191, 189, 165, 194, 167, 165, 173, 193, 171, 193, 168, 184, 166, 172, 187, 176, 187, 176, 165, 174, 194, 165, 182, 185, 195, 188, 176, 169, 178, 186, 182, 178, 167, 177, 174, 172, 168, 183, 188, 184, 176, 187, 173, 170, 189, 175, 184, 179, 186, 195, 192, 190, 185, 195, 173, 172, 194, 195, 171, 192, 172, 185, 195, 195, 190, 167, 190, 181, 188, 178, 172, 190, 181, 176, 194, 182, 181, 171, 186, 194, 174, 190, 169, 183, 165, 175, 179, 179, 189, 167, 178, 180, 192, 181, 168, 190, 187, 182, 176, 170, 171, 190, 193, 169, 177, 174, 182, 172, 193, 169, 189, 174, 166, 175, 188, 181, 181, 194, 170, 188, 171, 167, 182, 179, 169, 184, 179, 187, 187, 183, 172, 183, 186, 171, 181, 188, 172, 168, 183, 195, 185, 173, 166, 183, 182, 185, 193, 169, 168, 191, 177, 176, 169, 176, 187, 180, 180, 193, 171, 173, 174, 194, 172, 190, 176, 195, 189, 190, 186, 192, 187, 195, 176, 172, 181, 181, 171, 186, 185, 172, 185, 195, 181, 191, 172, 195, 188, 191, 183, 184, 187, 171, 174, 183, 182, 194, 173, 186, 193, 191, 192, 166, 172, 176, 174, 165, 166, 168, 186, 189, 175, 183, 174, 187, 190, 171, 167, 174, 194, 170, 195, 173, 191, 173, 172, 174, 166, 171, 176, 186, 188, 174, 169, 193, 167, 190, 190, 181, 191, 169, 185, 182, 178, 187, 195, 170, 183, 171, 186, 179, 193, 182, 172, 167, 179, 166, 167, 195, 192, 166, 171, 169, 178, 176, 168, 174, 183, 171, 194, 180, 166, 181, 172, 172, 180, 176, 180, 189, 171, 189, 168, 167, 181, 192, 188, 168, 165, 174, 187, 166, 177, 190, 175, 183, 172, 176, 175, 181, 185, 193, 173, 192, 166, 168, 185, 188, 167, 180, 180, 194, 194, 185, 171, 186, 166, 190, 171, 177, 
177, 170, 182, 165, 194, 190, 179, 178, 187, 186, 183, 173, 187, 168, 182, 168, 189, 183, 194, 169, 177, 169, 182, 186, 187, 174, 179, 184, 174, 195, 192, 178, 169, 176, 172, 195, 193, 180, 186, 166, 190, 172, 165, 195, 176, 179, 172, 175, 188, 180, 195, 190, 186, 177, 185, 175, 176, 181, 188, 181, 169, 190, 190, 179, 176, 188, 172, 194, 188, 174, 192, 165, 176, 175, 174, 175, 180, 184, 170, 165, 181, 184, 178, 174, 181, 165, 192, 175, 172, 182, 181, 165, 177, 189, 191, 181, 181, 174, 186, 170, 174, 183, 173, 191, 190, 166, 190, 177, 178, 178, 170, 195, 175, 189, 187, 171, 180, 173, 175, 185, 168, 168, 167, 182, 181, 187, 186, 175, 186, 173, 165, 193, 179, 188, 192, 191, 166, 194, 185, 183, 175, 177, 178, 179, 168, 173, 184, 190, 177, 165, 192, 183, 188, 172, 185, 178, 193, 190, 167, 178, 188, 168, 193, 175, 183, 187, 174, 174, 188, 190, 190, 172, 195, 188, 179, 187, 183, 180, 182, 174, 172, 189, 168, 172, 192, 186, 176, 166, 192, 186, 174, 191, 166, 187, 167, 176, 192, 168, 183, 166, 176, 167, 173, 188, 166, 189, 188, 190, 177, 168, 189, 170, 179, 187, 183, 191, 194, 191, 186, 180, 194, 185, 183, 180, 172, 169, 182, 182, 174, 183, 194, 172, 178, 178, 193, 166, 171, 188, 173, 187, 173, 177, 177, 183, 171, 191, 172, 190, 184, 172, 181, 181, 168, 176, 178, 168, 172, 166, 184, 186, 178, 172, 190, 181, 170, 192, 192, 180, 166, 188, 175, 189, 172, 171, 195, 190, 168, 178, 184, 194, 179, 184, 184, 170, 165, 176, 185, 182, 189, 186, 190, 195, 185, 178, 186, 174, 176, 181, 169, 189, 170, 185, 174, 185, 181, 190, 194, 167, 185, 179, 171, 170, 188, 171, 191, 189, 195, 195, 195, 173, 169, 174, 177, 186, 168, 175, 194, 171, 195, 191, 187, 191, 166, 173, 178, 169, 184, 190, 192, 173, 182, 185, 189, 189, 170, 166, 185, 171, 168, 181, 192, 168, 192, 181, 166, 179, 182, 169, 167, 174, 174, 171, 172, 174, 181, 167]
Y positions: [107, 104, 105, 105, 113, 103, 109, 113, 105, 114, 110, 114, 112, 110, 110, 112, 110, 103, 107, 103, 103, 111, 106, 101, 110, 107, 105, 112, 112, 115, 106, 106, 115, 107, 102, 103, 111, 110, 107, 105, 109, 107, 109, 114, 101, 101, 112, 115, 108, 111, 102, 110, 106, 110, 105, 107, 103, 103, 112, 112, 110, 116, 103, 101, 114, 111, 103, 110, 106, 103, 108, 101, 116, 116, 101, 109, 102, 101, 104, 103, 110, 115, 103, 116, 111, 116, 103, 115, 102, 115, 113, 102, 114, 108, 104, 105, 114, 103, 106, 111, 107, 113, 113, 101, 107, 108, 104, 116, 104, 110, 105, 116, 103, 114, 102, 101, 115, 106, 115, 113, 112, 105, 109, 115, 113, 111, 105, 104, 114, 103, 105, 115, 103, 108, 107, 114, 102, 116, 106, 115, 106, 112, 110, 106, 106, 102, 111, 115, 111, 106, 110, 112, 105, 105, 105, 114, 113, 101, 114, 109, 110, 101, 110, 113, 110, 107, 116, 115, 112, 101, 104, 114, 107, 115, 111, 108, 109, 106, 101, 115, 104, 106, 108, 114, 101, 101, 105, 104, 106, 114, 105, 112, 108, 116, 111, 105, 104, 115, 106, 106, 103, 101, 109, 108, 109, 106, 101, 102, 109, 102, 104, 114, 103, 110, 107, 110, 103, 116, 115, 114, 107, 108, 109, 109, 109, 115, 107, 115, 104, 111, 110, 112, 101, 110, 116, 103, 110, 103, 102, 110, 110, 113, 105, 112, 113, 110, 115, 107, 108, 105, 113, 110, 101, 112, 104, 116, 103, 107, 112, 106, 115, 111, 103, 113, 110, 104, 108, 106, 114, 113, 105, 101, 116, 112, 106, 102, 101, 109, 101, 116, 102, 102, 108, 103, 105, 104, 116, 103, 112, 111, 110, 109, 116, 103, 112, 110, 110, 116, 114, 112, 111, 106, 101, 110, 116, 107, 111, 110, 113, 102, 109, 112, 101, 114, 103, 116, 105, 116, 104, 113, 105, 101, 105, 106, 112, 116, 103, 116, 101, 103, 101, 106, 104, 105, 110, 105, 105, 112, 112, 108, 114, 106, 112, 114, 116, 116, 111, 114, 111, 109, 103, 108, 114, 112, 110, 111, 108, 103, 103, 106, 107, 102, 112, 104, 110, 116, 105, 112, 101, 104, 101, 108, 112, 106, 106, 102, 101, 115, 112, 106, 114, 106, 116, 103, 109, 103, 110, 105, 109, 113, 104, 113, 115, 116, 111, 102, 111, 101, 106, 107, 104, 105, 108, 101, 
108, 108, 116, 101, 107, 104, 106, 114, 110, 107, 115, 105, 107, 110, 103, 106, 109, 115, 112, 112, 109, 113, 106, 105, 102, 112, 110, 105, 104, 102, 109, 103, 108, 107, 115, 113, 111, 115, 113, 104, 110, 113, 115, 107, 103, 109, 112, 110, 105, 113, 102, 116, 115, 114, 102, 116, 112, 108, 109, 110, 111, 111, 102, 116, 102, 110, 113
\end{Verbatim}

\subsection{LLM answer and plan to Fig.~\ref{fig:main_snapshots} examples}
\label{sec:LLM_answer_and_plan}

\subsubsection{Coordination with Sonnet}

\begin{Verbatim}[breaklines=true]
I'll help you create a plan to effectively eliminate the enemies. Looking at the positions, I notice that all enemy units are currently in the Northern Forest (y-coordinates between 135-149), and they will need to move south through the middle of the map.

Let's analyze our forces:
- We have 500 spearmen (IDs 0-499)
- We have 500 archers (IDs 500-999)

Given the map's width (150) and our need to protect archers, I suggest creating three main battle groups across the middle of the map, each with both spearmen and archers. The spearmen will form a protective front line while archers attack from behind.

Here's the detailed plan:

BEGIN PLAN
Step 0:
prerequisites: []
objective: position
units: [0:167]
- target position: (25, 75)
- behavior: attack_in_close_range any
units: [167:334]
- target position: (75, 75)
- behavior: attack_in_close_range any
units: [334:500]
- target position: (125, 75)
- behavior: attack_in_close_range any
units: [500:667]
- target position: (25, 65)
- behavior: attack_in_long_range any
units: [667:834]
- target position: (75, 65)
- behavior: attack_in_long_range any
units: [834:1000]
- target position: (125, 65)
- behavior: attack_in_long_range any

Step 1:
prerequisites: [0]
objective: elimination all
units: [0:167]
- target position: (25, 75)
- behavior: attack_in_close_range any
units: [167:334]
- target position: (75, 75)
- behavior: attack_in_close_range any
units: [334:500]
- target position: (125, 75)
- behavior: attack_in_close_range any
units: [500:667]
- target position: (25, 65)
- behavior: attack_in_long_range any
units: [667:834]
- target position: (75, 65)
- behavior: attack_in_long_range any
units: [834:1000]
- target position: (125, 65)
- behavior: attack_in_long_range any
END PLAN

This plan creates three battle groups positioned across the map:
1. Western group (at x=25)
2. Central group (at x=75)
3. Eastern group (at x=125)

Each group has:
- About 167 spearmen in front (at y=75)
- About 167 archers behind them (at y=65)

The spearmen are positioned 10 units ahead of the archers to provide protection while still allowing the archers to shoot over them at approaching enemies.

The groups are spread out to cover the entire width of the map, ensuring no enemies can slip through. When enemies come down from the Northern Forest, they'll face a solid line of spearmen backed by archer fire.

Would you like me to adjust any aspect of this plan?
\end{Verbatim}

\subsubsection{Exploit weaknesses with Sonnet}
\begin{Verbatim}[breaklines=true]
Let me analyze the situation and the enemy positions.

The map is divided into 4 quadrants by two rivers that cross at (46-52, 47-53), with bridges allowing passage between these quadrants.

Enemy forces are organized in three distinct groups:
1. Northwest quadrant (around x:5-25, y:75-95): Mainly archers (health=2)
2. Northeast quadrant (around x:75-95, y:5-25): Mainly spearmen (health=24)
3. Northeast quadrant (around x:75-95, y:75-95): Mainly cavalry (health=12)

Our forces are currently positioned in the Southwest quadrant (around x:5-40, y:5-40).

Strategic considerations:
1. We should use cavalry against archers (NW group)
2. Archers against spearmen (NE group)
3. Spearmen against cavalry (NE group)

Here's my proposed plan:

BEGIN PLAN
Step 0:
prerequisites: []
objective: position
units: [0:250]
- target position: (19, 49)
- behavior: attack_and_move any
units: [250:500]
- target position: (21, 49)
- behavior: attack_and_move any
units: [500:750]
- target position: (20, 49)
- behavior: attack_and_move any

Step 1:
prerequisites: [0]
objective: position
units: [500:750]
- target position: (21, 76)
- behavior: attack_in_close_range archer
units: [250:500]
- target position: (76, 21)
- behavior: attack_in_long_range spearmen
units: [0:250]
- target position: (76, 76)
- behavior: attack_in_close_range cavalry

Step 2:
prerequisites: [1]
objective: elimination all
units: [500:750]
- target position: (15, 85)
- behavior: attack_in_close_range archer
units: [250:500]
- target position: (85, 15)
- behavior: attack_in_long_range spearmen
units: [0:250]
- target position: (85, 85)
- behavior: attack_in_close_range cavalry
END PLAN

Plan explanation:
1. First, we move all forces to the west bridge (Step 0)
2. Then split our forces (Step 1):
- Cavalry (500:750) goes north to engage archers
- Archers (250:500) goes east to engage spearmen
- Spearmen (0:250) goes northeast to engage cavalry
3. Finally, each group moves to eliminate their targeted enemies (Step 2)

This plan leverages each unit type's strengths against their natural enemies while using the bridges strategically to reach their positions.
\end{Verbatim}

\subsubsection{Markers with 4o-mini}
\begin{Verbatim}[breaklines=true]
To achieve your goal of moving the troops to the markers A, B, C, and D while ignoring enemies, we can create a plan that sequentially moves the units to each marker and then to the objective position at (61, 0).

However, since the objective is to reach the position at (61, 0) with at least one unit, we should ensure that the units are grouped effectively and can reach the objective after visiting the markers.

Let's break down the plan:

1. Move all units to marker A at (193, 85).
2. Move all units to marker B at (49, 136).
3. Move all units to marker C at (9, 134).
4. Move all units to marker D at (11, 9).
5. Finally, move all units to the objective position at (61, 0).

I will now create the detailed plan based on this sequence.

BEGIN PLAN
Step 0:
prerequisites: []
objective: position
units: all
- target position: (193, 85)
- behavior: follow_map

Step 1:
prerequisites: [0]
objective: position
units: all
- target position: (49, 136)
- behavior: follow_map

Step 2:
prerequisites: [1]
objective: position
units: all
- target position: (9, 134)
- behavior: follow_map

Step 3:
prerequisites: [2]
objective: position
units: all
- target position: (11, 9)
- behavior: follow_map

Step 4:
prerequisites: [3]
objective: position
units: all
- target position: (61, 0)
- behavior: follow_map
END PLAN

This plan allows your units to move to the specified markers and then to the objective position while ignoring enemies. Would you like to make any adjustments or proceed with this plan?
\end{Verbatim}

\subsubsection{Terrain with 4o}
\begin{Verbatim}[breaklines=true]
Here is the detailed plan based on your instructions:

BEGIN PLAN
Step 0:
prerequisites: []
objective: position
units: all
- target position: (164, 71)
- behavior: follow_map any

Step 1:
prerequisites: [0]
objective: position
units: all
- target position: (33, 159)
- behavior: follow_map any

Step 2:
prerequisites: [1]
objective: position
units: all
- target position: (11, 101)
- behavior: follow_map any

Step 3:
prerequisites: [2]
objective: position
units: all
- target position: (23, 44)
- behavior: follow_map any

Step 4:
prerequisites: [3]
objective: position
units: all
- target position: (61, 0)
- behavior: follow_map any
END PLAN
\end{Verbatim}

\subsubsection{Strategic points with o1-mini}
\begin{Verbatim}[breaklines=true]
BEGIN PLAN
Step 0:
prerequisites: []
objective: position
units: [0:39]
- target position: (45, 252)
- behavior: attack_in_close_range any
units: [350:389]
- target position: (45, 252)
- behavior: attack_in_long_range any

units: [39:78]
- target position: (138, 205)
- behavior: attack_in_close_range any
units: [389:428]
- target position: (138, 205)
- behavior: attack_in_long_range any

units: [78:117]
- target position: (205, 142)
- behavior: attack_in_close_range any
units: [428:467]
- target position: (205, 142)
- behavior: attack_in_long_range any

units: [117:156]
- target position: (252, 87)
- behavior: attack_in_close_range any
units: [467:506]
- target position: (252, 87)
- behavior: attack_in_long_range any

units: [156:195]
- target position: (272, 37)
- behavior: attack_in_close_range any
units: [506:545]
- target position: (272, 37)
- behavior: attack_in_long_range any

units: [195:234]
- target position: (29, 238)
- behavior: attack_in_close_range any
units: [545:584]
- target position: (29, 238)
- behavior: attack_in_long_range any

units: [234:273]
- target position: (65, 135)
- behavior: attack_in_close_range any
units: [584:623]
- target position: (65, 135)
- behavior: attack_in_long_range any

units: [273:312]
- target position: (135, 75)
- behavior: attack_in_close_range any
units: [623:662]
- target position: (135, 75)
- behavior: attack_in_long_range any

units: [312:350]
- target position: (249, 21)
- behavior: attack_in_close_range any
units: [662:700]
- target position: (249, 21)
- behavior: attack_in_long_range any
END PLAN

\end{Verbatim}

\subsection{Grammar for our Behavior trees}
\label{sec:BT_grammar}
\begin{Verbatim}[breaklines=true, breakautoindent=true]
    ?start: node

%import common.WS
%ignore WS

nodes : node ("::" node | "|>" node)*
node  :
    | sequence
    | fallback
    | action
    | condition

sequence  : "S" "(" nodes ")"
fallback  : "F" "(" nodes ")"
action    : "A" "(" atomic ")"
condition : "C" "(" atomic ")"

atomic :
    | move
    | attack
    | stand
    | follow_map
    | in_sight
    | in_reach
    | is_dying
    | is_armed
    | is_flock
    | is_type 
    | is_in_forest
    | success_action
    | failure_action
    

move      : "move" (direction | sense qualifier (foe | friend) (unit ("or" unit)* |any)?)
attack    : "attack" qualifier (unit ("or" unit)* |any)?
stand     : "stand"
in_sight  : "in_sight" (foe | friend) (unit ("or" unit)* |any)?
in_reach  : "in_reach" (foe | friend) source time (unit ("or" unit)* |any)?
is_dying  : "is_dying" (self | foe | friend) intensity
is_armed  : "is_armed" (self | foe | friend)
is_flock  : "is_flock" (foe | friend) direction
is_type   : "is_type" negation unit
follow_map : "follow_map" sense intensity?
is_in_forest : "is_in_forest" 
success_action : "success_action" 
failure_action: "failure_action"

sense     : /toward|away_from/
direction : /north | east | south | west | center/
foe       : /foe/
friend    : /friend/
qualifier : /strongest | weakest | closest | farthest | random/
intensity : /low | middle | high/
self      : /self/
unit      : /spearmen | archer| cavalry | balista | dragon | civilian/
any       : /any/
negation  : /a | not_a/
source    : /them_from_me | me_from_them/
time      : /now | low | middle | high/
\end{Verbatim}

\subsection{List of behavior trees used in this paper}
\subsubsection{Available for HIVE}
\label{sec:available_bts}

\begin{figure}[ht]
    \centering
    \includegraphics[width=\linewidth]{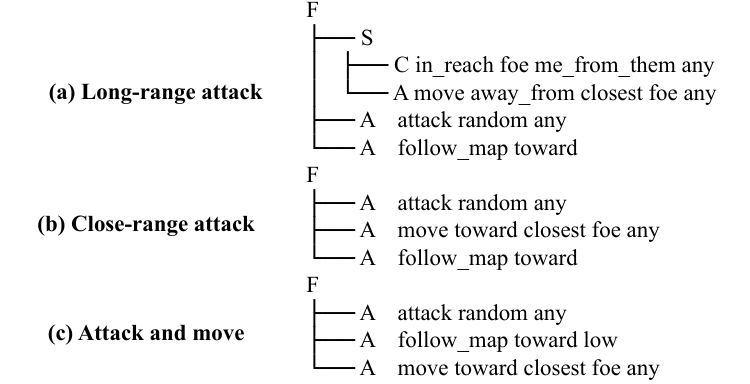}
    \caption{The three main behavior trees available to HIVE to control each unit.}
    \label{fig:the_three_bt}
\end{figure}
    
Long-range attack:
\begin{Verbatim}[breaklines=true, breakautoindent=true]
F(S(C( in_reach foe me_from_them high any) :: A (move away_from closest foe any)) :: A (attack random any) ::  A (follow_map toward))
\end{Verbatim}

Close-range attack:
\begin{Verbatim}[breaklines=true, breakautoindent=true]
F( A (attack random any) :: A (move toward closest foe any) :: A (follow_map toward))
\end{Verbatim}

Attack and move:
\begin{Verbatim}[breaklines=true, breakautoindent=true]
F( A (attack random any) :: A (follow_map toward low) :: A (move toward closest foe any) )
\end{Verbatim}

Move toward target position:
\begin{Verbatim}[breaklines=true, breakautoindent=true]
A (follow_map toward)
\end{Verbatim}

Stand:
\begin{Verbatim}[breaklines=true, breakautoindent=true]
A (stand)
\end{Verbatim}

\subsubsection{Only present in the opposing units}
The two following behavior trees are used by the opposing side in the Markers and Terrain ability test to avoid the enemy unit to get lost in the trees where they cannot see the player's units any longer.

Attack in long range and stay out of the forest:
\begin{Verbatim}[breaklines=true, breakautoindent=true]
F (S (C (is_in_forest) :: A (follow_map toward)) :: S(C( in_reach foe me_from_them high any) :: A (move away_from closest foe any)) :: A (attack closest any))
\end{Verbatim}

Attack in close range and stay out of the forest:
\begin{Verbatim}[breaklines=true, breakautoindent=true]
F (S (C (is_in_forest) :: A (follow_map toward)) :: A (attack closest any) :: A (move toward closest foe any))
\end{Verbatim}

\subsection{Prompts used for the 5 ability tests}
\label{sec:prompts_appendix_5_ability_tests}
\subsubsection{(a) Coordinate}
\begin{Verbatim}[breaklines=true, breakautoindent=true]
0. Make a plan that forms as many squads as you think necessary to cover the central row of the battlefield to eliminate all the enemies as fast as possible. Judisiously place our long range units so that they are not in close combat.
1. Design a plan that forms as many group of units as you think necessary to cover the central row of the battlefield to eliminate all the enemies as quickly as possible. Judisiously place our long range units so that they protected by our close range units.
2. Write down a plan that forms as many group of units as you think necessary to cover the central row of the map to eliminate all the enemies as fast as possible. Ensure to place our long range units so that they are not in close combat.
3. Design a plan that forms as many group of units as you think necessary to cover the central row of the map to eliminate all the enemies as quickly as possible. Make sure to place our long range units so that they in far range of the enemies.
4. Make a plan that forms as many squads as you think necessary to cover the middle row of the map to eliminate all the enemies as quickly as possible. Ensure to place our archers so that they protected by our close range units.
5. Design a plan that forms as many battalions as you think necessary to cover the middle row of the battlefield to eliminate all the enemies as fast as possible. Ensure to place our archers so that they in far range of the enemies.
6. Form as many battalions as you think necessary to cover the central row of the battlefield to eliminate all the enemies as fast as possible. Make sure to place our archers so that they are not in close combat.
7. Write down a plan that forms as many battalions as you think necessary to cover the central row of the battlefield to eliminate all the enemies as fast as possible. Make sure to place our long range units so that they protected by our close range units.
8. Make a plan that forms as many squads as you think necessary to cover the middle row of the battlefield to eliminate all the enemies as quickly as possible. Judisiously place our archers so that they in far range of the enemies.
9. Make a plan that forms as many group of units as you think necessary to cover the middle row of the battlefield to eliminate all the enemies as quickly as possible. Make sure to place our archers so that they in far range of the enemies.
\end{Verbatim}

\subsubsection{(b) Exploit weakness}
\begin{Verbatim}[breaklines=true, breakautoindent=true]
0. First, compute the positions and types of the enemy battalions. Then, design a plan that play on their weaknesses to split our units into three groups with specific strengths and send each one against one of the enemy squads to minimize our casualties.
 1. First, determine the positions and types of the enemy battalions. Then, make a plan that play on their weaknesses to split our army into three battalions with specific strengths and send each one against one of the enemy groups to maximize efficiency.
 2. First, analyse the situation to determine the positions and types of the enemy squads. Then, write a plan that play on their weaknesses to split our army into three battalions with specific strengths and send each one against one of the enemy battalions to minimize our casualties.
 3. First, analyse the situation to determine the positions and types of the enemy groups. Then, make a plan that play on their weaknesses to split our units into three squads with specific strengths and send each one against one of the enemy battalions to minimize our casualties.
 4. First, analyse the situation to compute the positions and types of the enemy squads. Then, design a plan that play on their weaknesses to split our army into three groups with specific strengths and send each one against one of the enemy battalions to minimize our casualties.
 5. First, analyse the situation to compute the positions and types of the enemy groups. Then, design a plan that play on their weaknesses to split our army into three squads with specific strengths and send each one against one of the enemy groups to maximize efficiency.
 6. First, analyse the situation to determine the positions and types of the enemy groups. Then, make a plan that play on their weaknesses to split our army into three squads with specific strengths and send each one against one of the enemy groups to minimize our casualties.
 7. First, analyse the situation to compute the positions and types of the enemy squads. Then, write a plan that play on their weaknesses to split our army into three groups with specific strengths and send each one against one of the enemy squads to maximize efficiency.
 8. First, compute the positions and types of the enemy squads. Then, make a plan that play on their weaknesses to split our army into three squads with specific strengths and send each one against one of the enemy battalions to maximize efficiency.
 9. First, compute the positions and types of the enemy battalions. Then, design a plan that play on their weaknesses to split our army into three squads with specific strengths and send each one against one of the enemy groups to minimize our casualties.
\end{Verbatim}

\subsubsection{(c) Follow markers}
The player mark four positions: 
\begin{itemize}
    \item A: (193, 85);
    \item B: (49, 136);
    \item C: (9, 134);
    \item D: (11, 9).
\end{itemize}

\begin{Verbatim}[breaklines=true, breakautoindent=true]
 0. Move the spearmen to each position in alphabetical order while ignoring enemies and then to the final objective position.
 1. Write a plan to move our troups to each marker  while ignoring enemies and then to the final objective position.
 2. Design a plan to move our spearmen to each marker in alphabetical order while ignoring enemies and then to the final objective position.
 3. Make a plan to move our spearmen to each position in alphabetical order while ignoring enemies and then to the final objective position.
 4. Write a plan to move our army to each position  while ignoring enemies and then to the final objective position.
 5. Write a plan to move my troups to the positions A, B, C, D while ignoring enemies and then to the  objective position.
 6. Move our troups to the markers A, B, C, D while ignoring enemies and then to the  objective position.
 7. Make a plan to move our army to the markers A, B, C, D while ignoring enemies and then to the  objective position.
 8. Move our spearmen to the positions ABCD while ignoring enemies and then to the  objective position.
 9. Make a plan to move the units to the markers A, B, C, D while ignoring enemies and then to the final objective position.
\end{Verbatim}

\subsubsection{(d) Exploit terrain}
\begin{Verbatim}[breaklines=true, breakautoindent=true]
0. Make a plan to move my spearmen inside the forest on the right  to the bridge, then make them rush to the forest on the west, slightly on top of the bridge . Then make them hide deep inside the forest on the west  of the map. Finally, make them rush south to the left-bottom corner of the map inside the forest and then to the objective, minimizing the time out of cover. Everything while ignoring enemies to not loose time fighting.
1. Write a plan to move my units inside the trees on the east  to the bridge, then make them rush to the forest on the west, slightly on top of the bridge . Then make them hide  inside the trees on the west  of the map. Finally, make them rush down to the south-east corner of the map inside the forest and then to the objective. Everything while ignoring enemies.
2. Write a plan to move my spearmen inside the trees on the right  to the bridge, then make them rush to the forest on the west, a bit north of the bridge as it is closer. Then make them hide deep inside the forest on the west most part of the map. Finally, make them rush  to the south-east corner of the map inside the forest and then to the objective, minimizing the time out of cover. Everything while ignoring enemies to not loose time fighting.
3. Make a plan to move our units inside the forest on the east not too far to the bridge, then make them rush to the trees on the west,  on top of the bridge to quickly regain cover. Then make them hide  inside the trees on the west  of the map. Finally, make them rush south to the left-bottom corner of the map inside the forest and then to the objective. Everything while ignoring enemies.
4. Design a plan to move our spearmen inside the trees on the east close to the bridge, then make them run to the forest on the left,  on top of the bridge to quickly regain cover. Then make them hide deep inside the forest on the left  of the map. Finally, make them rush south to the south-east corner of the map inside the forest and then to the objective. Everything while ignoring enemies.
5. Write a plan to move our army inside the forest on the right  to the bridge, then make them move to the forest on the west,  on top of the bridge as it is closer. Then make them hide deep inside the forest on the west most part of the map. Finally, make them rush  to the south-east corner of the map inside the forest and then to the objective, minimizing the time out of cover. Everything while ignoring enemies to not loose time fighting.
6. Write a plan to move my army inside the forest on the east not too far to the bridge, then make them rush to the forest on the west,  north of the bridge to quickly regain cover. Then make them hide deep inside the forest on the west most part of the map. Finally, make them rush down to the left-bottom corner of the map inside the forest and then to the objective. Everything while ignoring enemies to not loose time fighting.
7. Move the spearmen inside the trees on the east close to the bridge, then make them rush to the forest on the west, a bit on top of the bridge to quickly regain cover. Then make them hide  inside the trees on the left  of the map. Finally, make them rush  to the south-east corner of the map inside the forest and then to the objective. Everything while ignoring enemies to not loose time fighting.
8. Design a plan to move our spearmen inside the forest on the east close to the bridge, then make them move to the forest on the west, slightly on top of the bridge . Then make them hide deep inside the trees on the west most part of the map. Finally, make them rush  to the left-bottom corner of the map inside the forest and then to the objective. Everything while ignoring enemies.
9. Make a plan to move the army inside the forest on the right not too far to the bridge, then make them run to the trees on the west,  north of the bridge . Then make them hide  inside the trees on the left most part of the map. Finally, make them rush  to the south-east corner of the map inside the forest and then to the objective, minimizing the time out of cover. Everything while ignoring enemies.
\end{Verbatim}

\subsubsection{(e) Strategize points}
\begin{Verbatim}[breaklines=true, breakautoindent=true]
0. Split our units on all the bridges to defend them and make sure to send archers and spearmen units on each one.
1. Make a plan to split our army on all the bridges to guard them and make sure to send long-range and spearmen units on each one.
2. Design a plan to split our units on all the bridges to protect them and make sure to send long-range and spearmen units on each bridge.
3. Make a plan to split our units on all the bridges to guard them and make sure to send long-range and close-range units on each one.
4. Write a plan to split our army on all the bridges to protect them and make sure to send archers and close-range units on each bridge.
5. Make a plan to split our units on all the bridges to protect them and make sure to send archers and spearmen units on each one.
6. Split our army on all the bridges to guard them and make sure to send long-range and close-range units on each one.
7. Split our army on all the bridges to protect them and make sure to send archers and spearmen units on each one.
8. Design a plan to split our units on all the bridges to defend them and make sure to send archers and close-range units on each bridge.
9. Make a plan to split our army on all the bridges to guard them and make sure to send archers and close-range units on each bridge.
\end{Verbatim}

\subsection{Prompts used for HIVE alone}
\label{sec:HIVE_alone_appendix}

\begin{Verbatim}[breaklines=true, breakautoindent=true]
'First, study the situation to find a good approach to achieve this mission then design the corresponding plan.',
 'First, analyze the situation to find a good strategy to win this mission then write down the corresponding plan.',
 'First, study the situation to find a good approach to achieve this scenario then make the corresponding plan.',
 'First, analyze the situation to find a good strategy to win this mission then design the corresponding plan.',
 'First, study the situation to find a good approach to achieve this mission then write down the corresponding plan.',
 'First, analyze the situation to find a good strategy to achieve this mission then make the corresponding plan.',
 'First, analyze the situation to find a good strategy to achieve this scenario then write down the corresponding plan.',
 'First, analyze the situation to find a good approach to win this mission then make the corresponding plan.',
 'First, study the situation to find a good approach to win this scenario then make the corresponding plan.',
 'First, analyze the situation to find a good approach to achieve this scenario then design the corresponding plan.'
\end{Verbatim}

\end{document}